\documentclass[pdflatex,sn-mathphys-num]{sn-jnl}

\usepackage{graphicx}%
\usepackage{multirow}%
\usepackage{amsmath,amssymb,amsfonts}%
\usepackage{amsthm}%
\usepackage{mathrsfs}%
\usepackage[title]{appendix}%
\usepackage{xcolor}%
\usepackage{textcomp}%
\usepackage{manyfoot}%
\usepackage{booktabs}%
\usepackage{algorithm}%
\usepackage{algorithmicx}%
\usepackage{algpseudocode}%
\usepackage{listings}%
\usepackage[colorinlistoftodos]{todonotes}
\usepackage{caption}
\newcommand{\eg}{\textit{e.g.}}
\newcommand{\ie}{\textit{i.e.}}
\newcommand{\beginsupplement}{%
  \setcounter{figure}{0}%
  \setcounter{table}{0}%
  \captionsetup[figure]{name=Supplementary Figure}%
  \captionsetup[table]{name=Supplementary Table}%
}

\theoremstyle{thmstyleone}%
\theoremstyle{thmstyletwo}%

\theoremstyle{thmstylethree}%

\begin{document}

\title[Article Title]{Towards progressive whole-slide image analysis with multi-scale pyramidal network, a pathologist inspired approach}


\author*[1]{\fnm{Shuyang} \sur{Wu}}\email{frank.wu@ed.ac.uk}

\author[2]{\fnm{Yifu} \sur{Qiu}}\email{yifu.qiu@ed.ac.uk}
\author[3]{\fnm{John-Melle} \sur{Bokhorst}}\email{jmbokhorst@indicalab.com}
\author[3]{\fnm{Ines P} \sur{Nearchou}}\email{inearchou@indicalab.com}
\author[1]{\fnm{Sandrine} \sur{Prost}}\email{s.prost@ed.ac.uk}
\author[1]{\fnm{Jonathan A} \sur{Fallowfield}}\email{jonathan.fallowfield@ed.ac.uk}
\author[2]{\fnm{Hakan} \sur{Bilen}}\email{h.bilen@ed.ac.uk}
\author*[1]{\fnm{Timothy J} \sur{Kendall}}\email{tim.kendall@ed.ac.uk}

\affil[1]{\orgdiv{Institute for Regeneration and Repair}, \orgname{University of Edinburgh}, \orgaddress{\city{Edinburgh}, \postcode{EH16 4UU}, \country{UK}}}
\affil[2]{\orgdiv{School of Informatics}, \orgname{University of Edinburgh}, \orgaddress{\city{Edinburgh}, \postcode{EH8 9AB}, \country{UK}}}
\affil[3]{\orgname{Indica Labs, LCC}, \orgaddress{\street{8700 Education Pl NW}, \city{Albuquerque}, \postcode{87114}, \state{NM}, \country{USA}}}




\abstract{
Pathologists make a low power assessment followed by limited but targeted higher power evaluation of regions of interest as the standard approach in gold-standard clinical histopathological reporting. Multiple-instance Learning (MIL) is commonly used for slide-level predictions in computational pathology and approaches using multi-scale features have been developed. However, existing multi-scale MIL methods typically rely on inflexible multi-magnification inputs and computationally expensive architectures. We therefore conceptualise a more efficient MIL method that is explicitly designed to mimic the efficient gold-standard human pathologist approach to multi-scale assessment. In this paper, we propose the \textbf{Multi-scale Pyramidal Network (MSPN)}, a generalisable module for attention-based MIL. MSPN introduces progressive multi-scale whole-slide image analysis using only a single high-magnification input. It consists of (1) grid-based remapping that aggregates high-magnification features to derive spatially-aware coarse feature maps, and (2) the Coarse Guidance Network (CGN) that learns coarse contexts. We take MSPN as a generalisable module to 4 attention-based frameworks and benchmark on 11 clinically relevant tasks across 10 publicly available datasets with 2 foundation models. Our results demonstrate that by modelling the human pathologist paradigm in slide evaluation, MSPN consistently improves MIL across the compared configurations and is competitive on performance against 8 state-of-the-art multi-scale models while remaining lightweight and accessible in a manner that may increase implementability into under-resourced clinical systems.
}

\keywords{Machine Learning, Multiple-instance Learning, Multi-scale, Computational Pathology}



\maketitle

\section{Introduction}\label{intro}
Computational pathology (CPath) using multiple-instance learning (MIL) is now the standard formulation to solve many clinical prediction or classification tasks. MIL overcomes high computational costs imposed by gigapixel whole-slide images (WSIs) in a weakly-supervised manner that breaks down WSIs into small patches and allows the identification of the most salient instances in predicting the annotation label~\cite{ilse2018abmil}.

In routine clinical practice, pathologists assess tissue using objective lenses of varying magnification, usually from 1.25$\times$ to 40$\times$, allowing them to appreciate differing but equally important histological features~\cite{randell2012diagnosis}. High magnifications allow fine cellular features to be resolved whilst lower magnifications allow essential global tissue architecture to provide context~\cite{skrede2020crcprog, chen2022hipt}. The standard pattern of expert assessment incorporates both modes of evaluation by using an initial low-power evaluation to understand the global slide-level perspective that identifies discrete areas require subsequent higher power examination.

WSIs are usually pyramidal files which contain images acquired at multiple pre-determined resolutions defined by slide scanner hardware. The pyramidal multi-scale structure allows a pathologist reporting the case digitally to assess the tissue using the traditional pan-and-zoom approach used with a physical microscope and glass slide. In fact, this behaviour is so inherent in reporting workflows that some clinical grade reporting platforms track and record the area of the image assessed at each magnification (viewport tracking). 
 
Before MIL became the typical formulation, multi-scale information was explored with multiple convolutional neural networks (CNNs) used to process inputs from different scales~\cite{campanella2019cinlicalgrade,barbano2021unitopatho}. This demonstrated that information from different magnifications provided diverse and useful feature representations. Subsequently, approaches to the multi-scale composition of WSIs have been studied in the context of MIL applications to solve more challenging tasks such as biomarker classification, mutational signature prediction~\cite{gao2024her2, ElNahhas2025biomarker, myles2025surgen}, and prognosis prediction~\cite{chen2022pancancer, wang2023chief, jiang2024crc_prog, zhang2024nsclc_prog}.

In CPath, most weakly supervised slide-level analysis studies use single-scale images that tile patches at a high magnification equivalent to objective lenses of 20$\times$ or 40$\times$~\cite{lu_2021_data-efficient, chen2022pathomic, liang2023macronet}, while the additional utilisation of images at 5$\times$ and 10$\times$ magnification has only been incorporated in multi-scale studies. The most common of these approaches require real multi-scale inputs, patching each slide at rigid magnifications (\eg,\ 5$\times$, 10$\times$ and 20$\times$) and then handling the data through simple concatenation~\cite{marini2021concat, zhang2022multi-scale, liu-swetz2025multireso}, cross-scale attention~\cite{hashimoto2020multi-scale,deng2024cross-scale}, dual gated attention~\cite{thandiackal2022zoommil}, or integrated attention transformers~\cite{xiong2023hagmil}. The demand of these methods for multiple discrete input images complicates the data preprocessing, multiplies the cost of feature extraction, and fixes image use to manufacturer-predetermined magnification scales.

Alternative graph-based methods, whilst not true multi-scale, reveal local and global-level topological structures to allow context-aware learning~\cite{chen2021patchgcn,yu2022h2mil,fourkioti2024camil}. A graph-based enhanced localisation method using an attention-based framework was proposed to model local dependencies that can be incorporated with global representations~\cite{castro-mac2024sm}. However, these graph methods focus more on regional relationships between high magnification patches instead of explicitly introducing multi-scale learning, with the exception of H$^2$-MIL, which adopts a combined design that builds graph representations over rigid multi-scale inputs. 

Other transformer architectures have also been used in multi-scale models. For example, HIPT~\cite{chen2022hipt} uses a hierarchical vision transformer to extract patch tokens from large tiles for multi-scale learning and TransMIL~\cite{shao2021transmil} applies multi-scale convolutional kernels to process features under different granularity. These existing methods result in either inflexibility of multiple rigid inputs or complex and expensive architecture, and most are not analogous to the expert pathologist assessment paradigm of undertaking high-power assessment in areas of interest identified at low magnification with understanding a consequence of the integration of the assessments at differing magnification.

Such computational inefficiencies do not fit the current movement of the field to take advantage of foundation models (FMs) that now provide good feature representations to boost lightweight models~\cite{wang2021transpath, chen2024uni, vorontsov2024virchow, xu2024gigapath, lu2024conch, campanella2025benchmark}. Prior attempts to incorporate multi-scale context extracted from FMs have relied on knowledge distillation, aligning features extracted from high magnification patches with those from low magnification patches during FM pre-training or fine-tuning~\cite{chen2022hipt, tan2025pathme}. However, a generalisable, lightweight multi-scale module for MIL in the FM era has not yet been studied.

\begin{figure}[h]
\begin{center}
\centerline{\includegraphics[width=\textwidth]{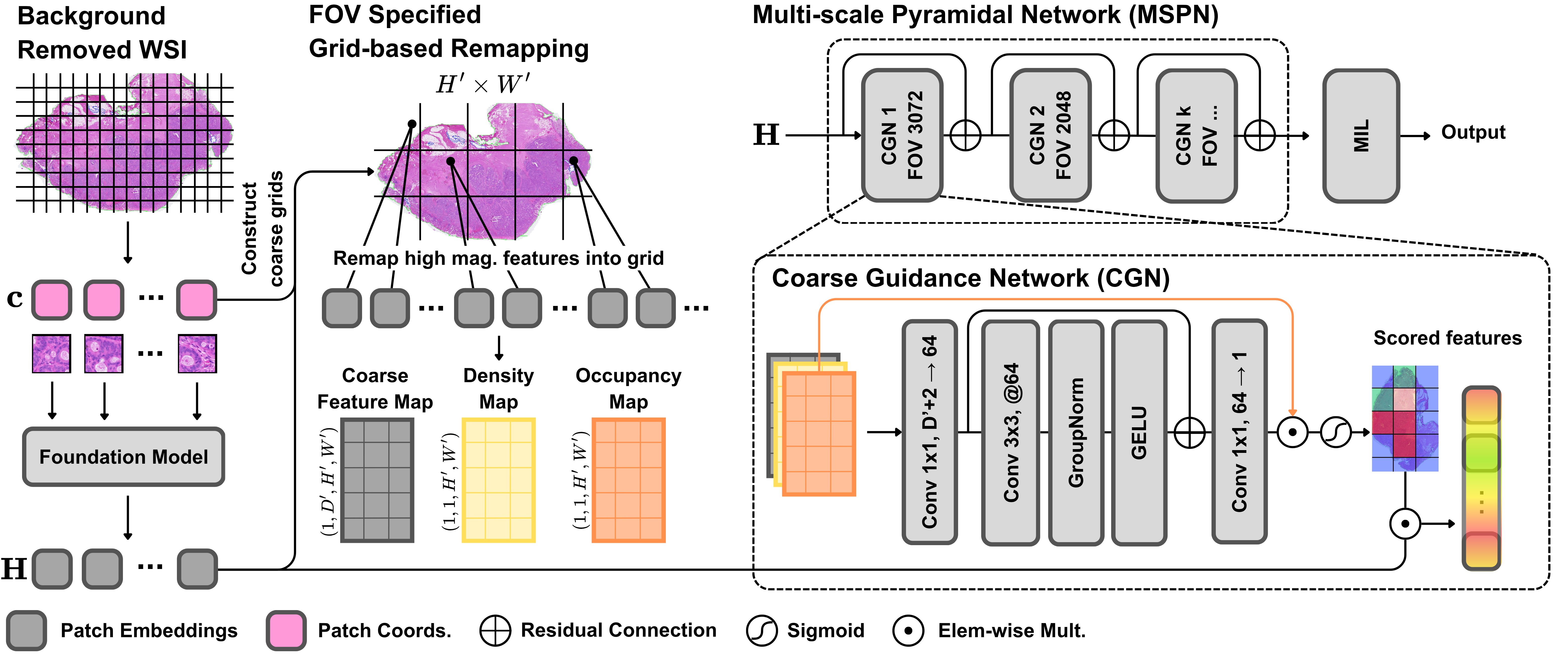}}
\caption{\textbf{Overview of the Multi-scale Pyramidal Network (MSPN) in a universal MIL pipeline.} (1) The MSPN consists of $k$ coarse guidance networks (CGN) with residual connections. Each CGN serves to generate guidances at coarser scales. (2) The coarse grids for guidances are determined by the selected field-of-view (FOV) using coordinates in the original high magnification, retaining the aspect ratio. (3) Features in each grid are aggregated to form a coarse feature map, a density map, and an occupancy map for the CGN. A sigmoid gate is learned to score the high magnification patches from the coarse feature map.}
\label{fig:overview}
\end{center}
\vskip -0.3in
\end{figure}

In this paper, we introduce the Multi-scale Pyramidal Network (MSPN), a lightweight, generalisable module that can be added to attention-based MIL frameworks to enable flexible and progressive multi-scale training (Figure~\ref{fig:overview}) in a manner informed by the human slide evaluation paradigm. We show that high magnification features from FMs can be directly aggregated to represent low magnification features, efficiently improving the performance of MIL downstream tasks without distillation and extra supervision. The MSPN consists of a sequence of residually connected Coarse Guidance Networks (CGNs) where each CGN serves the following functions: (1) creation of coarse-grids using high magnification coordinates, (2) remapping the patch-level features into each grid, (3) aggregation of features in each grid, and (4) generation of coarse guidances using convolution. The magnification of coarse guidances can be flexibly determined by modifying the grid size in each CGN. We demonstrate the efficacy of MSPN using 4 attention-based frameworks as main baselines (ABMIL~\cite{ilse2018abmil}, DSMIL~\cite{li2021dsmil}, CLAM-SB, and CLAM-MB~\cite{lu_2021_data-efficient}) by comparing different implementations: their vanilla implementations, multi-scale with concatenation, multi-scale with cross-scale attention (CSA)~\cite{deng2024cross-scale}, and multi-scale with MSPN. Performance is benchmarked on 11 tasks across 10 real-world datasets using features from 2 different pre-trained FMs. We show that through the explicit modelling of the human slide assessment paradigm, MSPN introduces significant performance improvements upon both the vanilla configuration and CSA and concatenation approaches to multi-scale inputs under the diverse combinations of MIL frameworks and FMs, and demonstrates competitive performance against the state-of-the-art multi-scale methods while staying lightweight and easy-to-use.

\section{Results}\label{sec:results}
We undertook benchmarking on 11 tasks across 10 real-world datasets using 2 different foundation models, including biomarker prediction in breast cancer, tumour histiotyping in renal cancer, and tumour grading in prostate cancer, described fully in Methods, to validate the efficacy of our MSPN.

\subsection{Accurate derivation of coarse features is possible through pooling of high magnification features}
\begin{figure}[h]
\begin{center}
\centerline{\includegraphics[width=\textwidth]{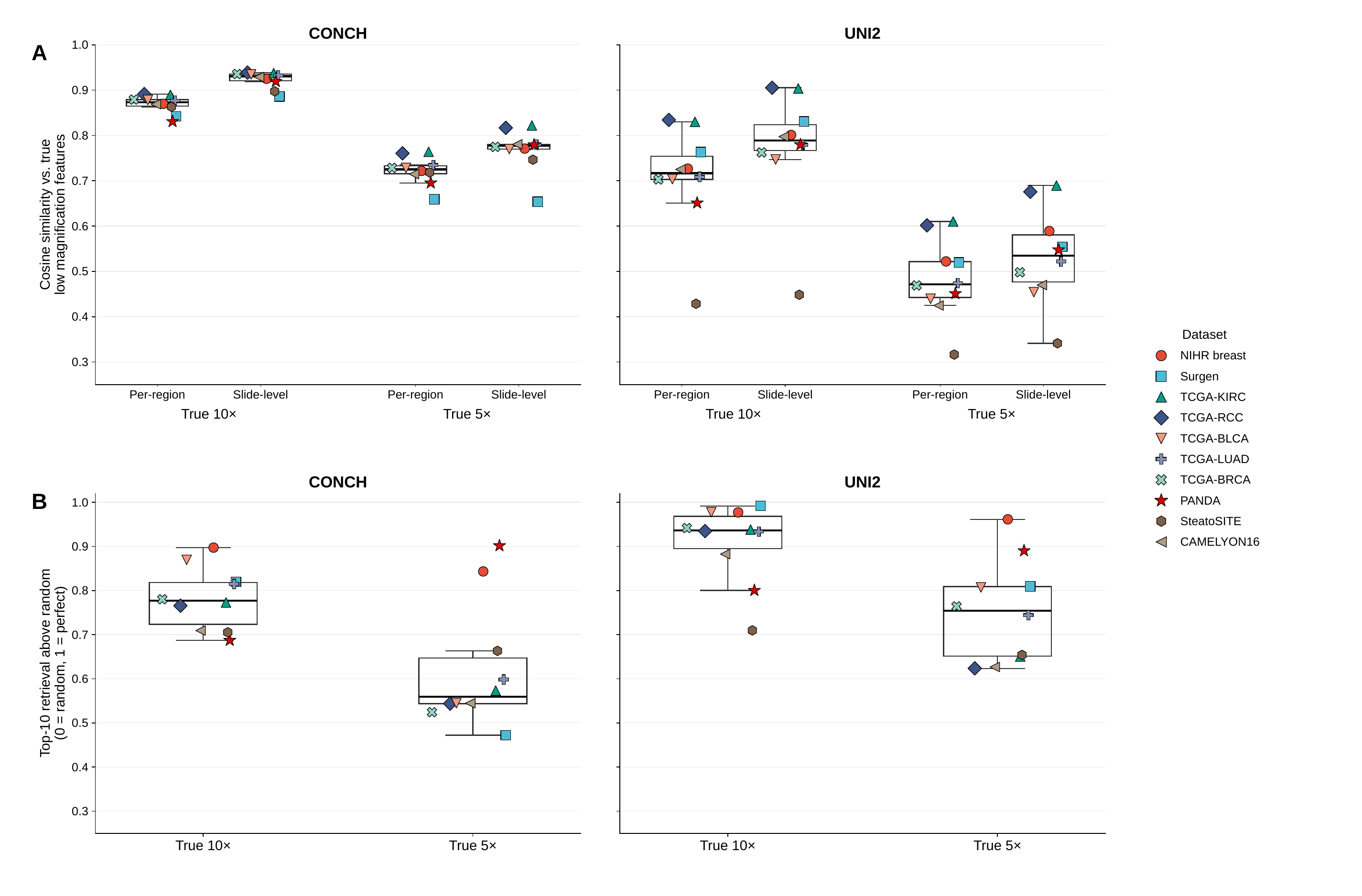}}
\caption{\textbf{Low magnification features can be accurately derived by meanpooling of $20\times$ magnification features.} (A) Cosine similarity of meanpooled derived features compared with the equivalent real low magnification features. (B) Rate of retrieval in the top 10 matching areas based on the derived low magnification features using their similarity with the real low magnification features. Retrieval above random is measured. All boxplots show median (centre line), first and third quartiles (lower and upper box limits), 1.5$\times$ IQR (whiskers).}
\label{fig:fidelity}
\end{center}
\vskip -0.3in
\end{figure}

We used mean-pooled high magnification features extracted by the FMs to derive low magnification features. To evaluate the similarity of low magnification feature sets derived from the high magnification input with the true low power features extracted by the FMs, we evaluated their cosine similarity, a mathematical measure of the similarity of two non-zero vectors calculated as the cosine of the angle between them, comparing those generated from $20\times$ magnification pooled features with the real lower magnification features across 10 datasets, 2 foundation models and 2 magnifications (Figure~\ref{fig:fidelity}). 

At a per-region level that measures the agreement within each grid region, the low-magnification ($5\times$ and $10\times$) feature sets derived from high-magnification features closely matched their real $5\times$ and $10\times$ counterparts, with average cosine similarities of 0.7225 and 0.8689, respectively, using CONCH as the foundation model, and 0.4828 and 0.7073 using UNI2. For slide-level features, derived by further mean-pooling all regions, showed even higher agreement with the real $5\times$ and $10\times$ features, with average cosine similarities of 0.7694 and 0.9235 for CONCH and 0.5341 and 0.7755 for UNI2.

We then measured the top-10 retrieval to determine if a specific region's low ($5\times$ and $10\times$) magnification feature sets derived from high magnification features can be ``picked out" from the set representing the whole WSI based on their similarity with the true low magnification features. Averaged over all datasets, true low-magnification features for a given region retrieved the pooled counterpart for the same region within the top-10 candidates reaches 79.1\% on CONCH and 91.5\% on UNI2 at 10$\times$ magnification, and 65.7\% (CONCH) and 78.7\% (UNI2) at 5$\times$ magnification. 

Combined, these data show that it is technically feasible to use high magnification features to reconstruct low magnification features without needing real low magnification inputs.

\subsection{MSPN models the human paradigm of ``low power screen, high power focus" slide assessment}
\begin{figure}[h]
\begin{center}
\centerline{\includegraphics[width=\textwidth]{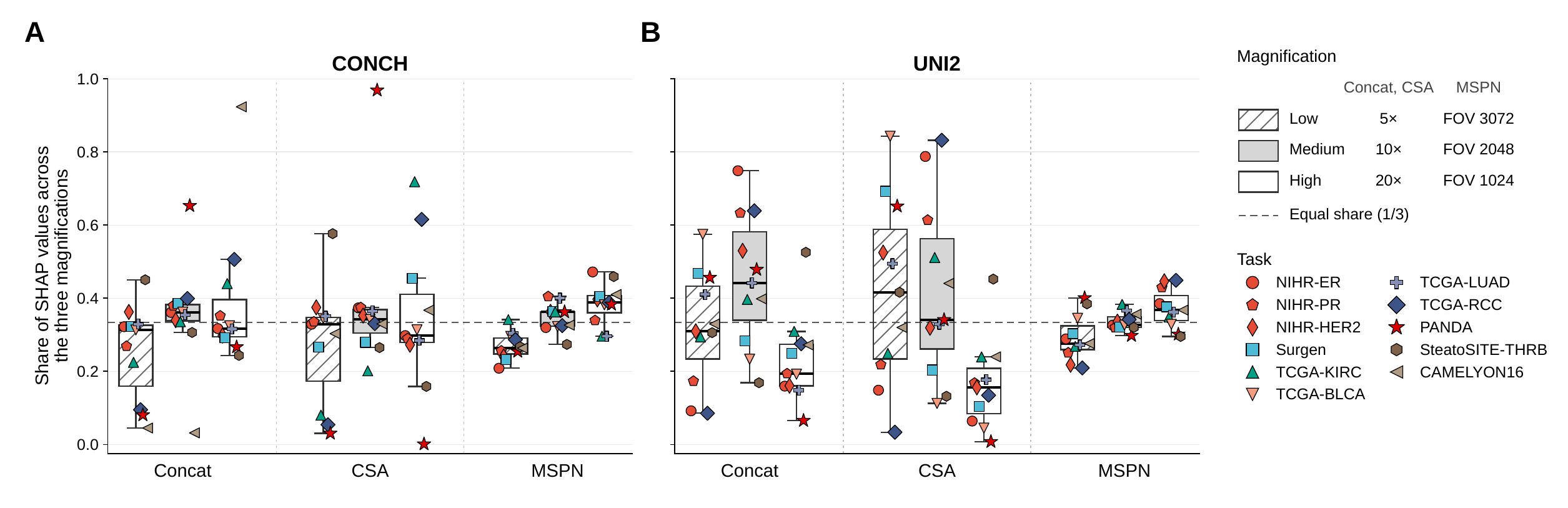}}
\caption{\textbf{SHAP analysis shows consistent use of magnifications by MSPN across all tasks.} SHAP analysis demonstrates the contribution/use of each magnification by the models in making slide-level predictions across all tasks, comparing ABMIL+Concat, ABMIL+CSA, and ABMIL+MSPN. In concatenation and cross-scale attention (CSA), one or two scales are often dominant, while for MSPN, each scale contributes consistently. All boxplots show median (centre line), first and third quartiles (lower and upper box limits), 1.5$\times$ IQR (whiskers).}
\label{fig:shap}
\end{center}
\vskip -0.3in
\end{figure}

\begin{figure}[h]
\begin{center}
\centerline{\includegraphics[width=\textwidth]{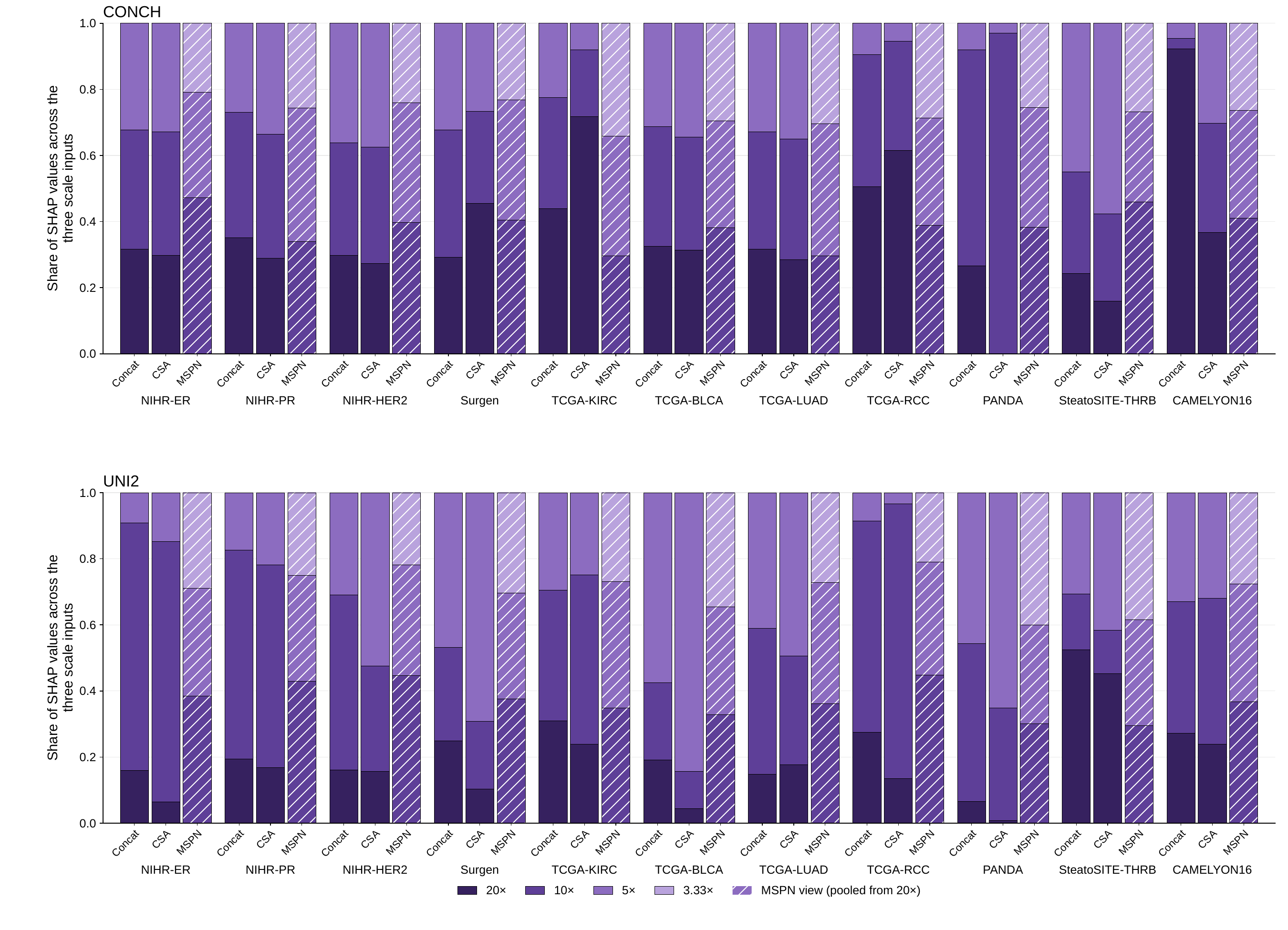}}
\caption{\textbf{Detailed utilisation of scales by the different methods of incorporating multi-scale inputs across all tasks.} The relative utilisation determined by SHapley Additive exPlanations analysis of each magnification by ABMIL with concatenation (Concat), cross-scale attention (CSA) and MSPN are shown for each of the 11 tasks using both CONCH and UNI2 as foundation  models.}
\label{supfig:shap_full}
\end{center}
\vskip -0.3in
\end{figure}

The human paradigm for slide assessment to take advantage of both architectural and cytological features within the tissue requires the consistent use of low, medium, and high magnification lenses. We performed SHapley Additive exPlanations (SHAP)~\cite{lungberg2017shap} analysis for predictions made by ABMIL+Concat, ABMIL+CSA and ABMIL+MSPN models across all tasks in Figure~\ref{fig:shap} and Supplementary Figure~\ref{supfig:shap_full}. SHAP offers explainability in machine learning by calculating the contribution of each input feature to the final prediction made by a model.

In multi-scale models using concatenation and cross-scale attention, there is inconsistency in the models' utilisation of the multi-scale inputs, with one or two scales often dominating. In contrast, MSPN uses inputs of each scale equally and consistently, computationally mimicking a pathologist's approach.

\begin{figure}[h]
\begin{center}
\centerline{\includegraphics[width=\textwidth]{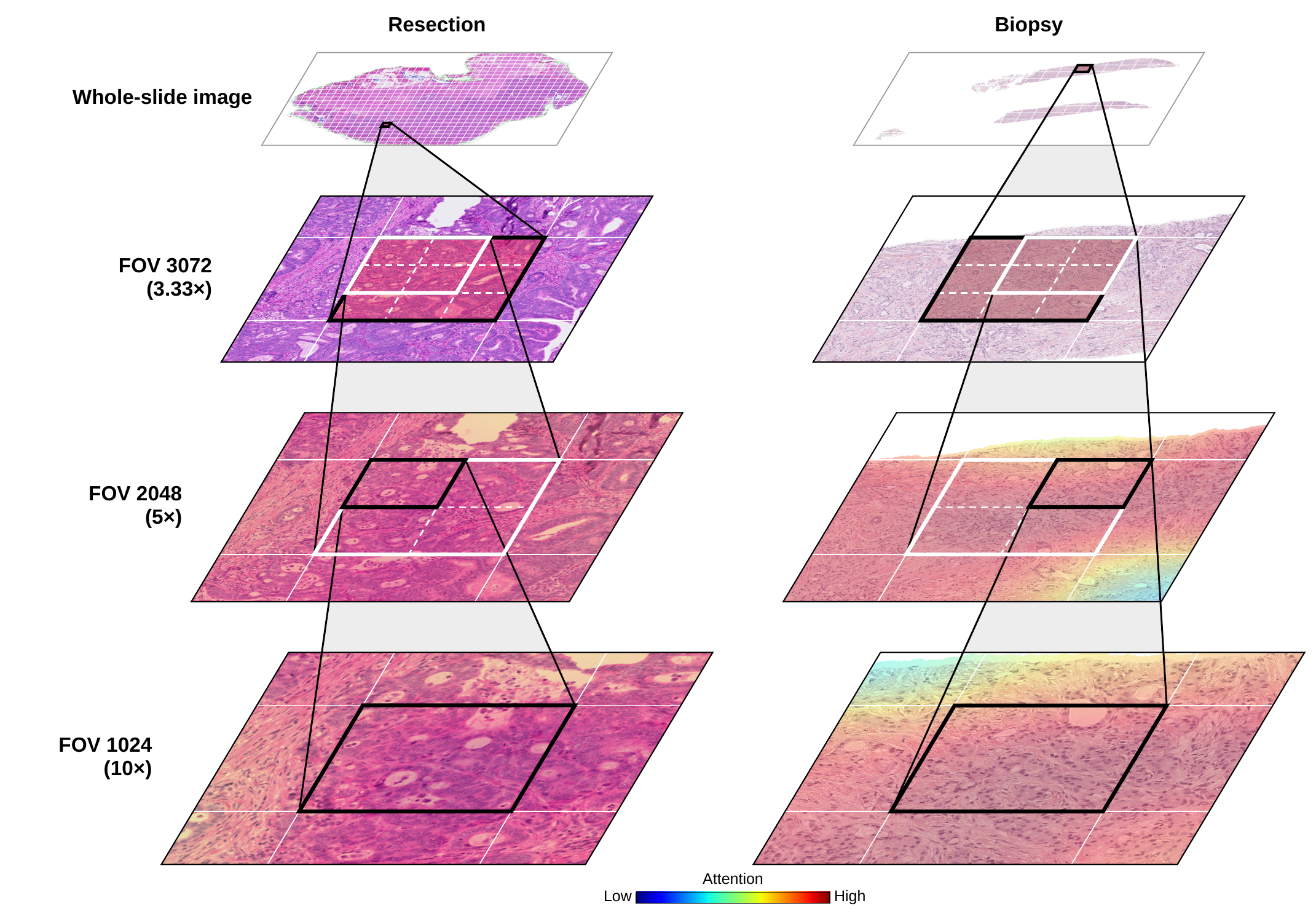}}
\caption{\textbf{A representation of the ``zoom-in" attention mechanism used by MSPN.} An H\&E-stained image of a colorectal cancer resection from the SurGen dataset (left) and an H\&E-stained image of a breast cancer biopsy from the NIHR dataset (right) are shown with magnification-level specific attention maps generated by the MSPN overlaid.}
\label{fig:zoom}
\end{center}
\vskip -0.3in
\end{figure}

Schematically, the attention map for each progressive magnification layer derived and progressively utilised during MSPN inference is shown in Figure~\ref{fig:zoom}. More fully, we visualised the attention heatmaps from the TCGA-BRCA external validation task, illustrated in Supplementary Figure~\ref{supfig:vis_simple}. A clear pattern of progressive refinement is observable, where the largest FOV of 3072 effectively captures broad, global regions of interest, creating a contextual map that guides the model. As the FOV decreases to 2048 and 1024, the attention progressively focuses on the more detailed morphological structures. The spatial alignment between the coarse guidances and the high-magnification features confirms that MSPN successfully learns from multi-scale contexts.

\subsection{MSPN enables performance improvements over existing multi-scale frameworks across diverse tasks}
We undertook task-specific comparison of MSPN versus the four attention-based baselines and their CSA and concatenation variants. MSPN demonstrates significant improvements over CSA and concatenation in 8 of 11 tasks using both CONCH and UNI2 as feature extractors (Figure~\ref{fig:mspn_vs_attn_radar}, detailed benchmarks shown in Supplementary Figure~\ref{supfig:mspn_vs_attn_bar}). MSPN variants show significant improvements in 136 of 264 comparisons, specifically on biomarker prediction of ER, PR, and HER2 in breast cancer, prognosis prediction in TCGA-KIRC and TCGA-LUAD datasets, subtyping of TCGA-RCC, Gleason grading of prostate cancer using the PANDA dataset, and tumour metastasis detection using the CAMELYON16 dataset. The 11-task summarised evaluation shows that MSPN outperformed the vanilla baselines and their CSA and concatenation multi-scale variants (Table~\ref{tab:model_level}, $p<0.001$).

\begin{figure}[h]
\begin{center}
\centerline{\includegraphics[width=\textwidth]{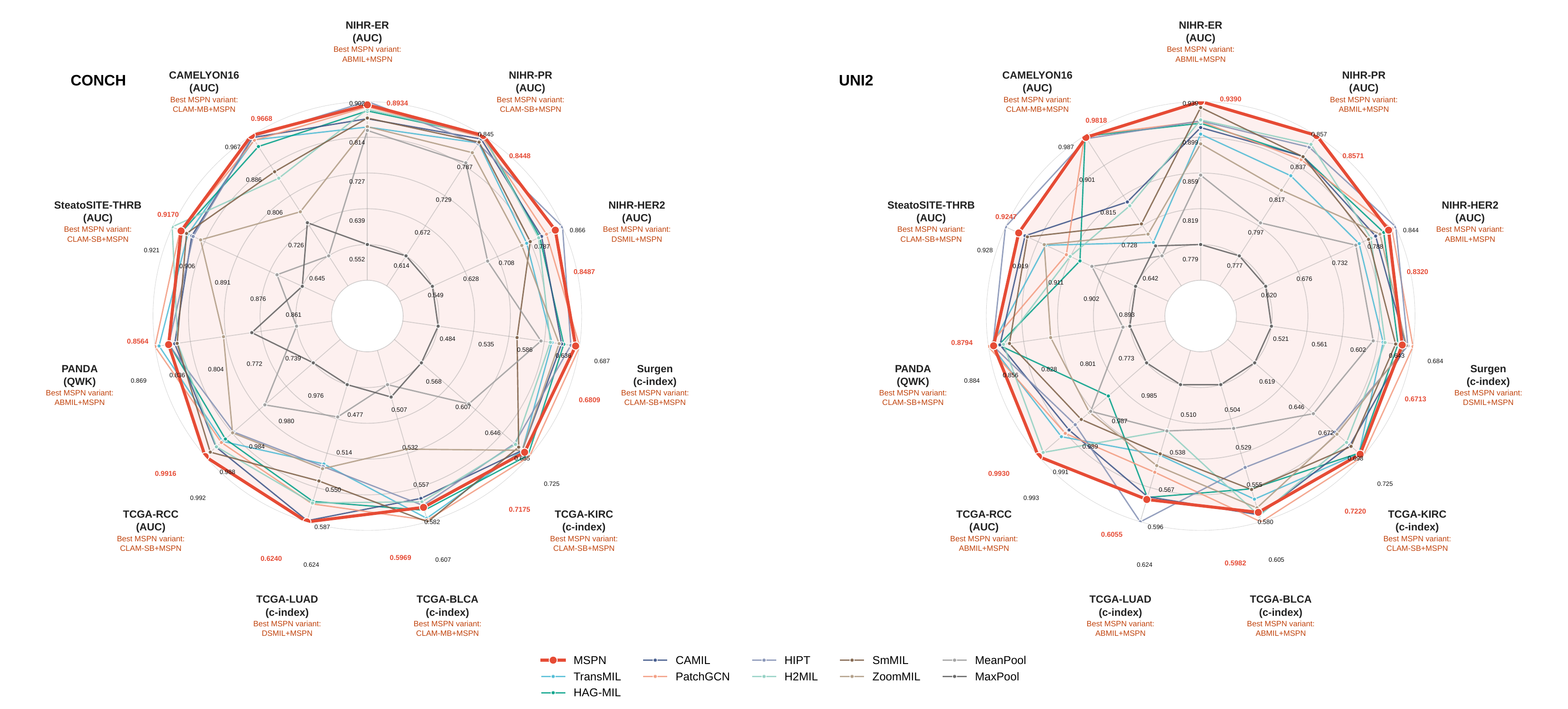}}
\caption{\textbf{MSPN matches or out-performs graph-based, transformer-based, and multi-magnification paradigms across all 11 tasks.} AUC score, c-index, and QWK are used to measure classification tasks, prognosis prediction tasks, and Gleason score grading, respectively. Mean values are indicated on the radial axes. The best performing MSPN variant in each task was selected for the comparisons.}
\label{fig:mspn_vs_others_radar}
\end{center}
\vskip -0.3in
\end{figure}

\begin{figure}[h]
\begin{center}
\centerline{\includegraphics[width=1.2\textwidth]{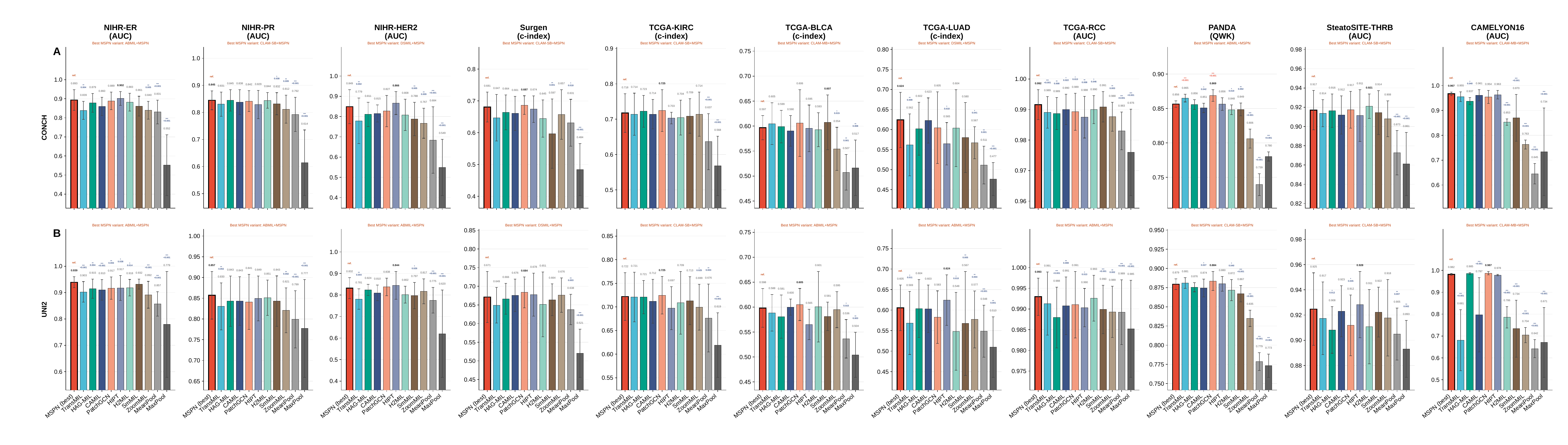}}
\caption{\textbf{Comparing MSPN with graph-based, transformer-based and multi-magnification paradigms across 11 tasks.} The results are reported in detailed. The best performing MSPN variant in each task is selected to compare. Mean values are reported and statistical tests are performed with 2000 paired bootstrap trails, taking the best MSPN variants as reference. Blue denotes significantly below the reference, while red denotes significantly above the reference. ***$p<0.001$, **$p<0.01$, *$p<0.05$.}
\label{supfig:mspn_vs_others_bar}
\end{center}
\vskip -0.3in
\end{figure}

We then compared MSPN with other baselines comprising graph-based, transformer-based, and multi-magnification designs (Figure~\ref{fig:mspn_vs_others_radar} and Supplementary Figure~\ref{supfig:mspn_vs_others_bar}). The best MSPN variant in each task significantly outperformed the baselines in 103 of 220 comparisons, with an average improvement of $0.0218$. The CLAM-SB+MSPN outperforms most baselines on TCGA-RCC across both UNI2 and CONCH foundation models and ABMIL+MSPN shows consistent improvement in NIHR-ER with UNI2. In comparison with the lightweight baselines, the best MSPN variants perform well on the NIHR-PR, NIHR-HER2, TCGA-LUAD, and CAMELYON16 tasks. Furthermore, MSPN outperforms larger scale models such as TransMIL, HIPT, and HAG-MIL in a variety of tasks.

Finally, to evaluate the generalisability of MSPN, we used TCGA-BRCA as an external validation cohort for biomarker prediction of ER ($n=1069$), PR ($n=1066$), and HER2 ($n=626$). MSPN demonstrates significant improvement in 37 of 72 comparisons (Table~\ref{tab:ext_roc}), with an average improvement of 0.0223 across four attention-based baselines and their CSA and concatenation multi-scale variants. Compared with the other baselines, MSPN significantly outperforms in 28 of 48 comparisons, with an average improvement in AUC scores of 0.0163, where the best MSPN reached AUC scores of 0.873 and 0.886 in ER, 0.793 and 0.799 in PR, and 0.670 and 0.687 in HER2, under CONCH and UNI2 features, respectively (Table~\ref{tab:ext_roc_baselines}).

\subsection{Multi-scale modelling using MSPN is computationally efficient}
The computational efficiency of MSPN is theoretically demonstrated in Section~\ref{sec:complexity}.


To further empirically demonstrate the performance-efficiency benefits of MSPN that follow from this theoretical explanation, we calculated the runtime of ABMIL+MSPN across various scales (FOV) settings and compared these against relative model performance across the 11 specimen tasks with that of ABMIL with multi-scale concatenation and CSA. The runtime of MSPN is well-controlled when resulting in performance improvements compared to concatenation and CSA.

\begin{figure}[h]
\begin{center}
\centerline{\includegraphics[width=\textwidth]{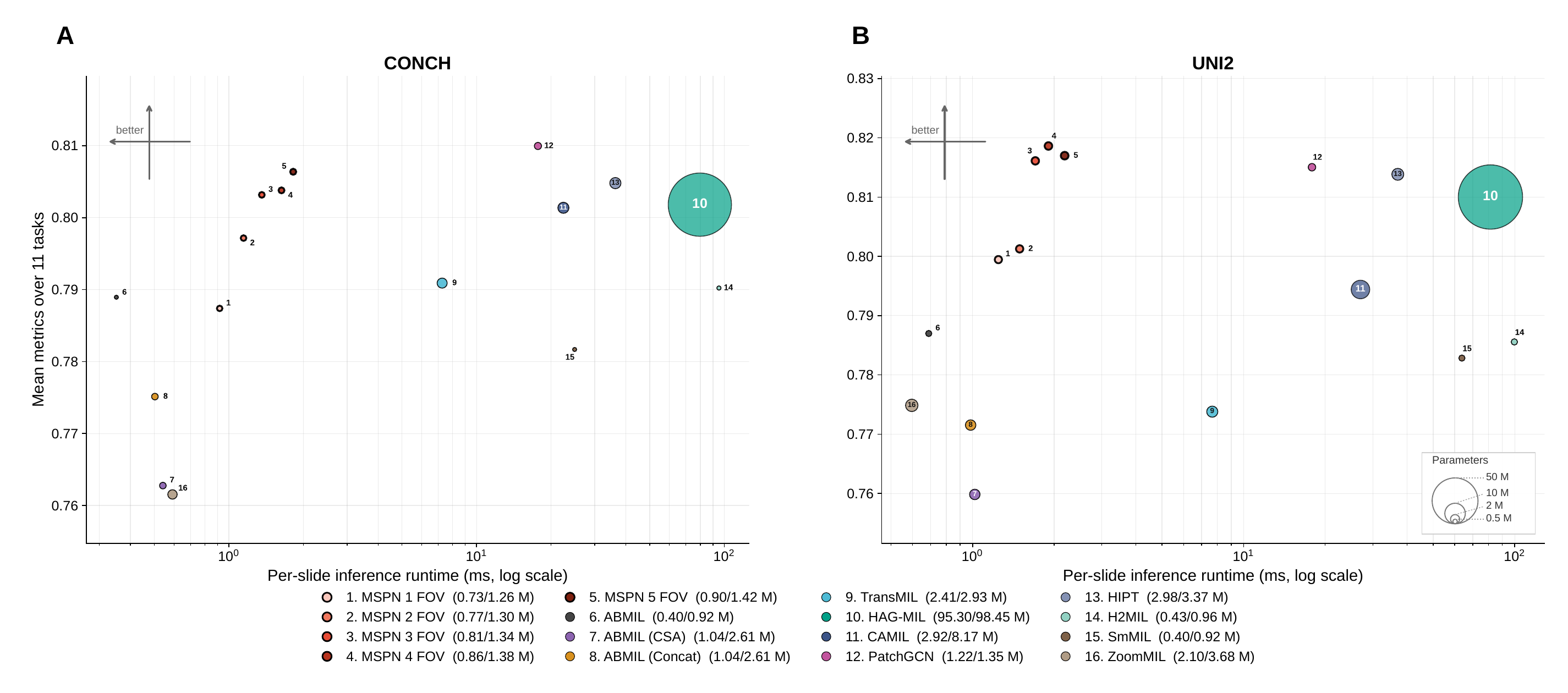}}
\caption{\textbf{MSPN maintains performance without excessive computational burden compared with other baselines across all tasks.} Model metrics, including AUC score, c-index, and QWK (measuring classification tasks, prognosis prediction tasks, and Gleason score grading, respectively), are averaged on 11 tasks. Runtime is measured n in milliseconds with median on 30 runs shown. Bubble size denotes parameter size, and the detailed parameter sizes of CONCH/UNI2 are shown in the legend. The patch number under each magnification is simulated based on the median number of patches in SurGen, $13,444$ for $20\times$, $3,422$ for $10\times$ and $884$ for $5\times$.}
\label{fig:efficiency}
\end{center}
\vskip -0.3in
\end{figure}

We also undertook a similar runtime-performance comparison with other published multi-scale models. While ZoomMIL is a more efficient baseline, its performance is limited. When compared with computationally heavier baselines, MSPN shows considerable improvements in efficiency while retaining accuracy. HAG-MIL uses a large parameter set for transformer blocks, yet its performance is similar to our low-cost approach. A similar pattern is evident with graph-based methods (e.g., CAMIL, Patch-GCN) which are also notably less efficient; H$^2$-MIL has a longer runtime because it builds graphs over multi-scale inputs (Figure~\ref{fig:efficiency}). MSPN remains lightweight and easy-to-use, operating at a small parameter size while delivering consistent improvements across tasks and foundation models.

\subsection{The inclusion of additional magnifications improves model performance}
\begin{figure}[h]
\begin{center}
\centerline{\includegraphics[width=\textwidth]{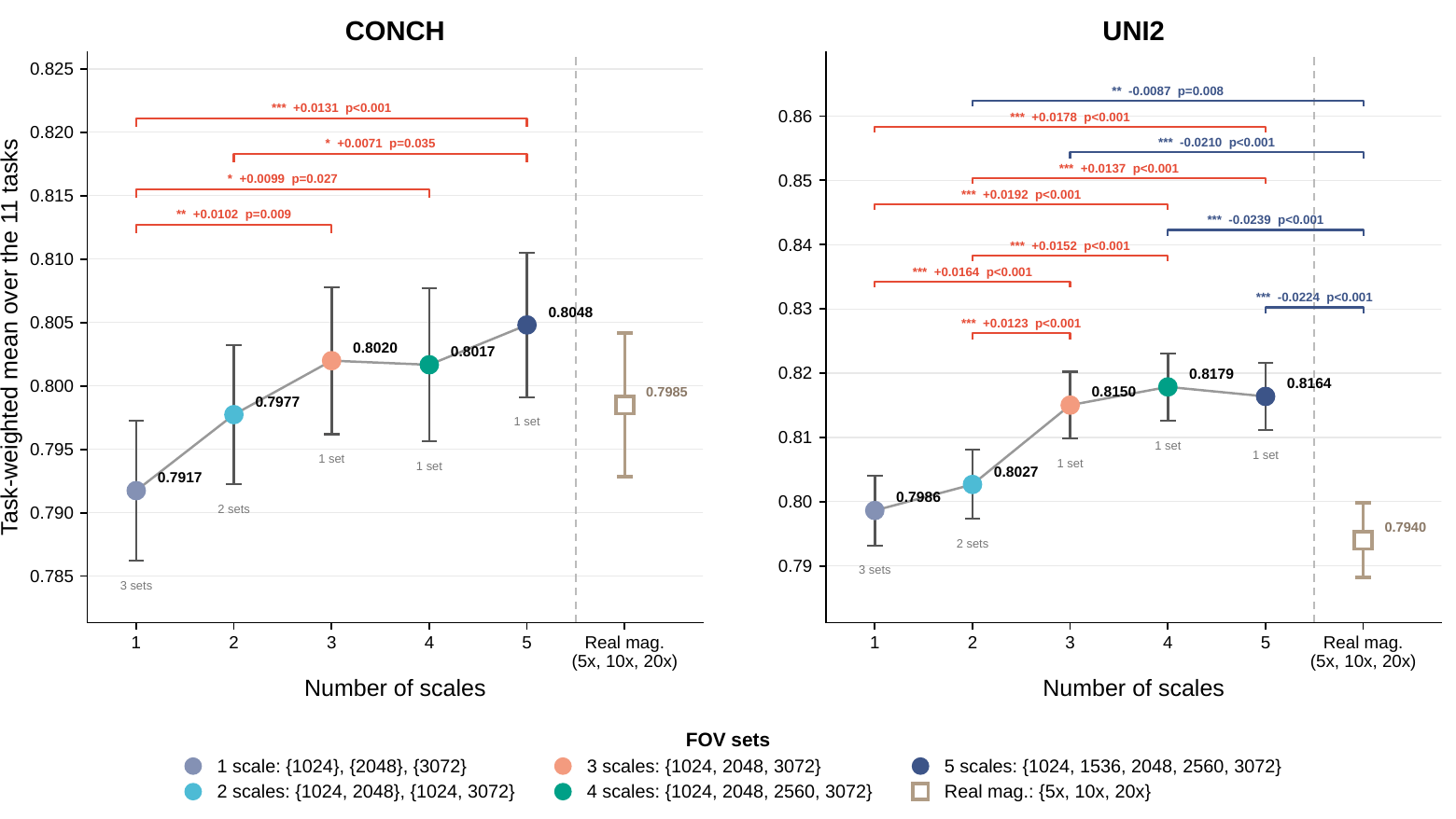}}
\caption{\textbf{MSPN with three levels of magnification offers the best performance across all tasks.} Model metrics, including AUC score, c-index, and QWK (measuring classification tasks, prognosis prediction tasks, and Gleason score grading, respectively) are averaged on 11 tasks. Standard deviation is reported and statistical tests are performed with 2000 paired bootstrap trails, taking the three-scale MSPN variant (\ie, FOV of $3072, 2048,1024$) as reference. ***$p<0.001$, **$p<0.01$, *$p<0.05$.}
\label{fig:abla_overall_diff_fov}
\end{center}
\vskip -0.3in
\end{figure}

Finally, we explored the performance of individual magnification (FOV) settings and their combinations to demonstrate the added value and optimal number of additional magnifications (Figure~\ref{fig:abla_overall_diff_fov}). Using features generated by UNI2, MSPN variants with more than one scale significantly outperformed their real magnification counterparts, further validating the use of high magnification features to derive lower magnification patterns. 

Performance improvements using both CONCH and UNI2 features were evident for configurations with three scales compared with their one- and two-scale counterparts. However, there was no significant further improvement using four- and five-scale variants compared with three-scale MSPN with FOVs of $3072, 2048$ and $1024$.

\section{Discussion}\label{discussion}
In this work, we aimed to model the human paradigm for slide evaluation where coarse architecture at low power is synthesised with finer microarchitectural and cytological features by the pathologist to derive a case level interpretation.  We conceived of a flexible, efficient and generalisable multi-scale learning model for MIL that are conceptually distinct from established multi-scale paradigms. Instead of relying on knowledge distillation, expensive transformer architectures, or rigid multi-scale patch extraction, MSPN spatially aggregates high magnification features to simulate lower magnification contexts on demand, following the formulation of MIL. In such a way, the number of scales and the field-of-view can be chosen by the user according to their needs and only a single high magnification feature set is extracted, which improves both flexibility and efficiency of multi-scale MIL. Further, the architecture of MSPN ensures that training and inference use information from multiple scales that capture the histological features in the manner similar to a pathologist's assessment.

The comparison against the published baselines focuses on the performance-efficiency trade-off, since attention-based frameworks are valued for their lightweight nature, and we demonstrate that MSPN can strengthen lightweight model performance without significant efficiency cost. Graph-based methods model relationships between high magnification patches to capture regional context but do not introduce an explicit scale hierarchy, so the gains we observe from adding progressively coarser guidance are complementary to the graph methods. TransMIL~\cite{shao2021transmil} utilises multi-scale convolutional kernels for pyramid positional encoding at different resolutions, but coerces features into square feature maps in a manner that discards the true aspect ratio, and hence actual tissue architecture, of the slide, whereas our grid-based remapping preserves it by reconstruction. As reported in Section~\ref{sec:results}, the substantially larger transformer and graph-based baselines do not convert their parameter size and inference speed into a corresponding accuracy advantage compared with our lightweight module.

Finally, our results demonstrate that multi-scale context can be obtained from foundation model features through a simple pooling operation. Previous approaches align high and low magnification representations with distillation during pre-training or fine-tuning~\cite{chen2022hipt, tan2025pathme}, which couples the multi-scale capability to the encoder and requires it to be retrained. We find that a frozen high magnification encoder already carries enough information for coarse context to be derived by spatial aggregation alone, with no distillation objective and no additional supervision. This decouples multi-scale learning from the encoder and places it in the downstream module, where it costs little and can be flexibly applied to any foundation model.



\section{Methods}
\subsection{Tasks and Datasets}
For robust evaluation and proof of generalisability, we use 11 clinical tasks on 10 real-world datasets.

\bmhead{Tumour classification}
Identification of tumour-containing regions is a common task in CPath. We employ CAMELYON16~\cite{camelyon16}, a breast cancer lymph node dataset that contains a total of 399 WSIs, comprising 239 cases without metastatic tumour and 160 cases containing metastasis. The slides are publicly available through Grand Challenge CAMELYON16.

\bmhead{Tumour subtyping}
To examine tumour subtype classification, we use the TCGA Renal Cell Carcinoma (RCC) dataset for subtyping, which contains a total of 919 WSIs, composed of 107 kidney chromophobe (KICH), 517 kidney renal clear cell carcinoma (KIRC), and 295 kidney renal papillary cell carcinoma (KIRP).

\bmhead{Biomarker prediction}
For tissue biomarker prediction tasks, we use on breast cancer and steatotic liver disease. The breast cancer receptor status prediction of oestrogen receptor (ER), progesterone receptor (PR) and human epidermal growth factor receptor 2 (HER2) are performed on the NIHR Breast Cancer dataset, which contains 491 WSIs. The prediction of receptor status is clinically important that informs treatment decision making. The ER and PR annotations are binary according to the Allred scores~\cite{iccr_allred}, where 0 and 2 are negative while 3-8 to be positive. For HER2 status prediction, 0 and 1+ are negative, 3+ are positive, while the borderline 2+ cases are further tested with fluorescence in situ hybridisation (FISH) and FISH amplified cases are positive. 

ER, PR and HER2 status prediction is further externally validated on The Cancer Genome Atlas (TCGA) Breast Invasive Carcinoma (BRCA) datasets that contains over 1,000 WSIs. After removing WSIs without required annotations, TCGA-BRCA has 1,069 cases for ER prediction, 1,066 cases for PR prediction and 626 cases for HER2 prediction.

Liver thyroid hormone receptor beta (THRB) prediction is performed on the SteatoSITE~\cite{kendall2023steatosite} dataset that contains 508 WSIs with associated RNAseq. THRB regulon activity defined from bulk RNAseq was negatively associated with clinical outcomes in the pan-Scotland Metabolic dysfunction-Associated Steatotic Liver Disease (MASLD) data commons, and Resmetirom, a THRB agonist, has been shown to improve outcomes in patients with the progressive form of MASLD~\cite{polyzos2025thrb, liu2026thrb}. The dataset contains 320 THRB regulon low and 188 THRB regulon high cases.

\bmhead{Tumour grading}
The Prostate cANcer graDe Assessment (PANDA) dataset of 10,433 WSIs was used for a tumour grading task. Prostate carcinoma is assigned a Gleason grade and represents a six-class classification problem. Cases in the dataset are annotated according to International Society of Urological Pathology (ISUP) grade group system, with scores from 0 to 5. The ISUP grade group is determined by the combination of the first and second most common Gleason patterns. We formulated the task to evaluate a model's ability to quantify the percentage of each pattern. 

\bmhead{Prognosis prediction}
Cancer prognosis is important for patients yet challenging to predict directly from stained tissue. We perform four prognosis prediction tasks using colorectal cancer, kidney renal clear cell carcinoma, urothelial bladder carcinoma, and lung adenocarcinoma datasets.

SurGen~\cite{myles2025surgen} is a publicly available dataset containing H\&E-stained whole-slide images (WSIs) from 843 colorectal cancer (CRC) resection cases. We take 425 cases that have follow-up survival data for prognosis prediction. The TCGA-KIRC dataset in the most aggresive subset within the TCGA-RCC dataset and contains 517 WSIs with 170 events and 345 censored cases. The TCGA-urothelial bladder carcinoma (BLCA) contains 452 WSIs, with 210 events and 242 censored cases. The TCGA-lung adenocarcinoma (LUAD) contains 527 WSIs, with 206 events and 321 censored cases.

\subsection{Preprocessing}
The preprocessing protocol is consistent across all datasets, without data curation or normalisation. WSIs with the highest magnification of $40\times$ are standardised to 0.2631 microns per pixel (MPP), while WSIs with highest magnification of $20\times$ are standardised to 0.5 MPP. Patches from three scales of each dataset are obtained, under the magnification of $5\times$, $10\times$ and $20\times$, where $20\times$ is high magnification and the other two are lower magnifications. We remove background, downsample to the desired magnification (\ie, $5\times$,$10\times$ and $20\times$) and tile patches with a non-overlapping size of $256\times 256$ pixels using the CLAM toolkit~\cite{lu_2021_data-efficient} and OpenSlide~\cite{goode2013openslide}.

The tiled patches are then encoded with a feature encoder to obtain the slide-level embeddings. We evaluate our method by taking two specialised foundation models as encoders, CONCH~\cite{lu2024conch} and UNI2~\cite{chen2024uni}. CONCH outputs instance features of size $1\times 512$, while the instance feature size is $1\times 1536$ for UNI2.

\subsection{Problem formulation}
MIL is commonly formulated as a binary classification problem. Given a bag with $N$ instances that $X=\{x_1,x_2,...,x_N\}$, MIL is trained to predict the bag-level label $Y\in \{0,1\}$ with instance-level labels $\{y_1,y_2,...y_N\}$ unknown.
\begin{equation}
Y =  
    \begin{cases}
    0, & \text{iff} \sum_{n} y_{n}=0, \\
    1, & \text{otherwise.}
    \end{cases}
\end{equation}
Practically, MIL can further be extended to handle multi-class output in tasks such as subtyping~\cite{shao2021transmil,javed2022additive} and prognosis prediction~\cite{chen2022pathomic, chen2022pancancer}.

The current CPath pipeline typically includes two steps: (1) frozen feature extraction and (2) downstream task training. During feature extraction, a foundation model $g$ is used to extract features $\mathbf{h}_n$ for each instance, where $\mathbf{h}_n=g(x_n)\in \mathbb{R}^{1\times D}$ with $D$ denotes the feature dimension. The instance-level features are concatenated to form the bag-level feature $\mathbf{H}=\{\mathbf{h}_1,\mathbf{h}_2,...\mathbf{h}_n\}\in \mathbb{R}^{N\times D}$. In downstream task training, MIL is applied by aggregating the bag-level features into a bag representation for classification that $\varphi(f(\mathbf{H}))$, where $f$ denotes the MIL aggregator and $\varphi$ is a classifier with fully connected layers. Our key innovation is to enable flexible and progressive multi-scale learning, by adding our multi-scale proposal network between these two steps to create the multi-scale bag-level feature representation.

\subsection{FOV-specified Grid-based Remapping}
The aim for grid-based remapping is to avoid the need for multiple inputs, and instead utilise the high magnification features $\mathbf{H}$ to derive a coarse feature map that retains the spatial and contextual information. The extent of the tissue and the patch coordinates are used to construct coarser grids, and the features falling inside each grid cell are aggregated to form coarse-level representations. 

For a WSI of the height and width of $H$ and $W$ containing a bag of patches with coordinates $\mathbf{c}=(\mathbf{x}, \mathbf{y})\in\mathbb{R}^{N\times2}$ and $x_n\in [0,W],y_n\in[0,H]$, grid-based remapping is applied to create coarse feature maps using the high magnification bag-level feature $\mathbf{H}$ and the normalised patch coordinates $\mathbf{c'}=(\mathbf{x'},\mathbf{y')}\in \mathbb{R}^{N\times 2}$ that $x_n',y_n'\in[0,1]$. Firstly, a field-of-view (FOV) $s$ is selected to define the pseudo grid. For example, an $s=1024$ mimics the coarser tile size of $1024\times1024$, and a grid with $H'\times W'$ where $H'=\lceil\frac{H}{s}\rceil$ and $W'=\lceil\frac{W}{s}\rceil$ is created, preserving the WSI's aspect ratio. The grid coordinates are denoted as $(\mathbf{x^g},\mathbf{y^g}),x_n^g\in[0,W'],y_n^g\in[0,H']$. Subsequently, each $\mathbf{h}_n$ is remapped onto the grid according to their coordinates. In this step, we index the grid region in a row-major order, which builds a sequence of $\mathbf{idx}\in\mathbb{R}^N$ that maps each high magnification patch to its related coarser grid. If a grid includes no patches, zero-padding is applied. Next, intra-grid aggregation is applied by meanpooling the patches assigned to each region to obtain the coarse feature map $\mathbf{M}\in\mathbb{R}^{1\times D\times H'\times W'}$. As tissue is distributed in a WSI irregularly, many boundary regions are empty. To mitigate the influence of these zero feature vectors, we make the sampling explicit by appending two scalar channels per cell. Let $\mathbf{C}$ to be a matrix recording patch counts in each region, where $\mathbf{C}\in\mathbb{Z}^{1\times1\times H'\times W'}$ and $C_{max}$ denotes the maximum patch counts across regions, a binary occupancy mask is written as $\mathbf{O}=\mathbf{1}[\mathbf{C}>0]\in \{0,1\}^{1\times 1\times H'\times W'}$ to filter out empty grids, and a density is written as $\mathbf{D}=\mathbf{C}/\max(C_{max},1)\in [0,1]^{1\times 1\times H'\times W'}$ to give the relative number of patches contributing to each grid.

The theoretical soundness of grid-based remapping is derived from the formulation of MIL. Given the simplest scenario where meanpooling is adopted as the MIL aggregator, the bag-level representation $\mathbf{b}$ can be written as $\mathbf{b}=f_{mean}(\mathbf{H})\in \mathbb{R}^{1\times D}$. Considering the $m^{th}$ grid as a small bag $X_m'=\{x_1,x_2,...,x_J\}$, where $J$ denotes the number of instances in the grid and $J<N$, the features in the grid are $\mathbf{H}_m'=\{\mathbf{h}_1,\mathbf{h}_2,...,\mathbf{h}_J\}\in \mathbb{R}^{J\times D}$. Hence, the representation of the small bag over meanpooling is $\mathbf{b}_m'=f_{mean}(\mathbf{H}_m')$. Following this, the coarse representation of the whole bag can be rewritten as $\mathbf{H}_{coarse}=\{\mathbf{b}_1',\mathbf{b}_2',...,\mathbf{b}_M'\}\in\mathbb{R}^{M\times D}$, where $M=H'\times W'$. Therefore, the grid-based remapping maintains the MIL formulation, but downsampled the bag-level features into a coarser scale that enables convolutional operation. The pseudo code of grid-based remapping is described in Algorithm~\ref{alg:gbr}.

\begin{algorithm}[h]
  \caption{Grid-based Remapping}
  \label{alg:gbr}
  \begin{algorithmic}
    \State {\bfseries Input:} features $\mathbf{H}\in\mathbb{R}^{N\times D}$, coords $(\mathbf{x},\mathbf{y})$, field-of-view $s$
    \State {\bfseries Output:} coarse feature map $\mathbf{M}$, occupancy $\mathbf{O}$, density $\mathbf{D}$, patch index $\mathbf{idx}$
    \State 1) Create grids: $W'\leftarrow\max(1,\lceil W/s\rceil)$, $H'\leftarrow\max(1,\lceil H/s\rceil)$, $M\leftarrow H'W'$
    \State 2) Normalise coords: $\mathbf{x'}\leftarrow(\mathbf{x}-x_{\min})/W$, $\mathbf{y'}\leftarrow(\mathbf{y}-y_{\min})/H$, where $x_n',y_n'\in[0,1]$
    \State 3) Give patch index:
    \State $\mathbf{u}\leftarrow\min(\lfloor\mathbf{x'}\cdot W'\rfloor,\,W'-1)$, $\mathbf{v}\leftarrow\min(\lfloor\mathbf{y'}\cdot H'\rfloor,\,H'-1)$
    \State $\mathbf{idx}\leftarrow\mathbf{v}\cdot W'+\mathbf{u}$, where $\mathbf{v}\cdot W'$ indexes the row and $\mathbf{u}$ indexes the column, $\mathbf{idx}\in\mathbb{Z}^{N}$
    \State 4) Remapping and intra-grid aggregation:
    \State $\mathbf{M}_{0}\leftarrow\text{torch.zeros}(M,D)$, $\mathbf{C}_{0}\leftarrow\text{torch.zeros}(M)$ \# initialise
    \State $\mathbf{M}_{sum}\leftarrow\mathbf{M}_{0}.\text{index\_add}(0,\mathbf{idx},\mathbf{H})$ \# mapping into grid
    \State $\mathbf{C}\leftarrow\mathbf{C}_{0}.\text{index\_add}(0,\mathbf{idx},\text{torch.ones}(N))$ \# patch count per grid
    \State $\mathbf{M}\leftarrow\text{reshape}(\mathbf{M}_{sum}/\mathbf{C}.\text{clamp}(\min{=}1).\text{unsqueeze}(1))$ \# average on each grid
    \State 5) Occupancy and density:
    \State $\mathbf{O}\leftarrow\text{reshape}(\mathbf{C}>0)$, $\mathbf{D}\leftarrow\text{reshape}(\mathbf{C}/\max(C_{max},1))$
  \end{algorithmic}
\end{algorithm}

\subsection{Coarse Guidance Network}
The coarse guidance network (GCN) is a lightweight convolutional network for coarser-level feature learning that aims to provide guidance over the coarse areas of interest at the selected scale, while also avoiding the high complexity of transformers and self-attention. The CGN processes the concatenated feature map $\mathbf{\bar{M}}=[\mathbf{M},\mathbf{O},\mathbf{D}]\in\mathbb{R}^{1\times (D'+2)\times H'\times W'}$, in which the pooled cell features have first been projected from $D$ to the hidden channel $D'$, with convolutional layers and outputs the coarse guidance $\mathbf{P}\in \mathbb{R}^{1\times 1\times H'\times W'}$. We construct CGN with three convolutional layers, where the first uses Conv$_{1\times 1}$ to fuse the occupancy and density channels into the features, the second uses Conv$_{3\times 3}$ with padding of 1 followed by group normalisation and a GELU activation and is applied as a residual branch, and the last uses Conv$_{1\times 1}$ whose output is masked by $\mathbf{O}$ so that grids containing no tissue provide no guidance, and the hidden channel $D'$ is set to $64$ in the implementation to limit computational complexity.

The guidance $\mathbf{P}$ is then used to generate a patch-wise attention map $\mathbf{M}_A\in \mathbb{R}^{1\times N}$ by assigning the grid value in $\mathbf{P}$ to each included patch, according to the patch index from the previous step of grid-based remapping. This functions to generate a coarse guidance for the whole high magnification slide. Next, element-wise multiplication is applied to score the slide-level features, obtaining the scored features $\mathbf{H}_k$ where $k$ denotes the features output by the $k^{th}$ CGN. The implementation of a single CGN is shown in Algorithm~\ref{alg:cgn}.

\begin{algorithm}[h]
  \caption{Coarse Guidance Network}
  \label{alg:cgn}
  \begin{algorithmic}
    \State {\bfseries Input:} data $\mathbf{H}$, coords $(\mathbf{x},\mathbf{y})$, field-of-view $s$
    \State {\bfseries Output:} scored feature $\mathbf{H}_{k}$
    \State $\mathbf{M},\ \mathbf{O},\ \mathbf{D},\ \mathbf{idx}\leftarrow\text{grid\_based\_remapping}(\mathbf{H},(\mathbf{x},\mathbf{y}),s)$
    \State $\mathbf{\bar{M}}\leftarrow\text{concat}([\,\text{proj}_{D\rightarrow D'}(\mathbf{M}),\ \mathbf{O},\ \mathbf{D}\,])$ \# $\mathbf{\bar{M}}\in\mathbb{R}^{1\times(D'+2)\times H'\times W'}$
    \State $\mathbf{\bar{M}}_1\leftarrow\text{Conv}_{1\times1}(\mathbf{\bar{M}})$ \# fuse occupancy and density
    \State $\mathbf{\bar{M}}_2\leftarrow\mathbf{\bar{M}}_1+\text{GELU}(\text{GroupNorm}(\text{Conv}_{3\times3}(\mathbf{\bar{M}}_1)))$ \# residual branch
    \State $\mathbf{P}\leftarrow\text{Conv}_{1\times1}(\mathbf{\bar{M}}_2)\odot\mathbf{O}$ \# empty grids give no guidance
    \State $\mathbf{M}_A\leftarrow\text{flatten}(\mathbf{P})[\mathbf{idx}]$,\ where $\mathbf{P}_{\text{flatten}}\in\mathbb{R}^{H'\cdot W'}$, $\mathbf{idx}\in\mathbb{Z}^{N}$ and $\mathbf{M}_A\in\mathbb{R}^{N}$
    \State $\mathbf{H}_k\leftarrow\mathbf{H}\odot\sigma(\mathbf{M}_A)$
  \end{algorithmic}
\end{algorithm}

\subsection{Multi-scale Pyramidal Network}
To enable flexible multi-scale learning and also share information across scales, we design the multi-scale pyramidal network (MSPN) that simply integrates multiple CGNs using residual connections; the overview of MSPN is shown in Figure~\ref{fig:overview}. The MSPN is built with $k$ blocks of CGNs, where coarse guidances are generated progressively from large to small FOV. Subsequently, the multi-scale aggregated output $\mathbf{H}_{mspn}$ can directly be used by any attention-based MIL framework as an input for downstream tasks. The pseudo-code of the forward process of MSPN is described in Algorithm~\ref{alg:mspn}. Denoting $\hat{Y}$ as the prediction and $\mathbf{s}$ as selected FOVs, an example of ABMIL~\cite{ilse2018abmil} with MSPN plugged-in is written as:
\begin{equation}
    \hat{Y}=\text{ABMIL}(\text{MSPN}(\mathbf{H},(\mathbf{x},\mathbf{y}),\mathbf{s}))
\end{equation}

Under this scheme, MSPN is designed to be optimised along with the task-specific loss to ensure generalisability and applicability to different tasks and architectures.

\begin{algorithm}[h]
  \caption{Multi-scale Pyramidal Network}
  \label{alg:mspn}
  \begin{algorithmic}
    \State {\bfseries Input:} data $\mathbf{H}$, coords $(\mathbf{x},\mathbf{y})$, field-of-views $\mathbf{s}=\{s_1,\dots,s_K\}$ sorted $s_1>\dots>s_K$
    \State {\bfseries Output:} multi-scale aggregated features $\mathbf{H}_{mspn}$
    \For{$k=1$ {\bfseries to} $K$}\Comment{coarse to fine}
    \State $\mathbf{H}_k\leftarrow \text{CGN}_k(\mathbf{H},(\mathbf{x},\mathbf{y}), s_k)$
    \State $\mathbf{H}\leftarrow\mathbf{H}+\mathbf{H}_k$
    \EndFor
    \State $\mathbf{H}_{mspn}\leftarrow\mathbf{H}$
  \end{algorithmic}
\end{algorithm}

\subsection{Computational Complexity of MSPN}
\label{sec:complexity}
The efficiency of the use of high magnification features to derive features representing other scales can be described theoretically. Let $M$ denote the number of grids $H'\times W'$. A single CGN performs three kinds of work, namely (1) remapping $N$ patch features into the grid and obtaining the coarse features, taking $\mathcal{O}(ND)$; (2) projecting the coarse features from $D$ to $D'$, taking $\mathcal{O}(MDD')$; and (3) the convolutions over the coarse features, taking $\mathcal{O}(MD'^2)$. The complexity of a single CGN is therefore:
\begin{equation}
\mathcal{T}_\text{CGN}=\mathcal{O}(ND+M(DD'+D'^2))\\
\end{equation}
, where a large hidden channel $D'$ results in squared complexity. Hence, $D'$ is set to be $64$ in the implementation, as stated in the previous section.

Next, denoting $k$ as the number of fields-of-view chosen (\ie, the number of CGNs), the complexity of an MSPN module consisting of $k$ CGNs is computed as:
\begin{equation}
    \mathcal{T}_\text{MSPN}=\mathcal{O}(kND+\sum_{i=1}^k M_i(DD'+D'^2))\\
\end{equation}
Therefore, the complexity of MSPN is linear with the patch number $N$ and the number of CGNs $K$, which keeps the cost well controlled under a large patch number and a varying number of CGNs, so long as $D'$ remains small.

\subsection{Learning objectives}
We take MSPN with ABMIL as an example to show the loss functions. For tumour classification, biomarker prediction, tumour subtyping, and tumour grading, MSPN is optimised with the common cross-entropy loss.
\begin{equation}
\hat{Y}=\text{ABMIL(MSPN}(\mathbf{H}, (\mathbf{x},\mathbf{y}),\mathbf{s}))
\end{equation}
\begin{equation}
\mathcal{L}_{cls} = CELoss(\hat{Y}, Y)
\end{equation}

Under the formulation of prognosis prediction, patients with a recorded death are considered as uncensored ($c=0$), while patients alive at last follow-up are censored ($c=1$). $\alpha$ is a variable for down-weighting the censored loss. Let $T$ denote the number of discrete time bins, $y\in\{1,\dots,T\}$ the observed bin of a patient, and $Y_{hazard}(t)$ and $Y_{surv}(t)$ the predicted hazard and survival rate at bin $t$, respectively; the censored loss $\mathcal{L}_{censored}$, uncensored loss $\mathcal{L}_{uncensored}$ and the loss for prognosis prediction $\mathcal{L}_{surv}$ are defined as below and we set $T=4$ and $\alpha=0.15$ following previous work~\cite{chen2022pancancer, zadeh2021nllsurv}:
\begin{equation}
Y_{hazard}=\text{Sigmoid}\big(\text{ABMIL}(\text{MSPN}(\mathbf{H},(\mathbf{x},\mathbf{y}),\mathbf{s}))\big)\in\mathbb{R}^{T}
\end{equation}
\begin{equation}
Y_{surv}(t)=\prod_{\tau\le t}\big(1-Y_{hazard}(\tau)\big),\qquad Y_{surv}(0)=1
\end{equation}
\begin{equation}
\mathcal{L}_{censored}=-\,c\,\log\big(Y_{surv}(y)\big)
\end{equation}
\begin{equation}
\mathcal{L}_{uncensored}=-\,(1-c)\Big[\log\big(Y_{surv}(y-1)\big)+\log\big(Y_{hazard}(y)\big)\Big]
\end{equation}
\begin{equation}
\mathcal{L}_{surv}=(1-\alpha)\big(\mathcal{L}_{censored}+\mathcal{L}_{uncensored}\big)+\alpha\,\mathcal{L}_{uncensored}
\end{equation}
, where an uncensored patient survived up to bin $y$ and then died within it, giving the likelihood $Y_{surv}(y-1)\,Y_{hazard}(y)$, whereas a censored patient is only known to have survived through bin $y$, giving $Y_{surv}(y)$.

\subsection{Experimental details}
Multi-scale MILs are usually implemented with three scales (\ie, $5\times$, $10\times$ and $20\times$). Therefore, we built MSPN with three FOVs (\ie, 3072, 2048 and 1024, in pixels, which correspond to $3.33\times$, $5\times$ and $10\times$) for comparison.

To rigorously benchmark the contribution of MSPN, we take ABMIL~\cite{ilse2018abmil}, DSMIL~\cite{li2021dsmil}, CLAM-SB, and CLAM-MB~\cite{lu_2021_data-efficient} as the attention-based platforms, and study three different types of multi-scale learning on them, including (1) concatenation, (2) cross-scale attention~\cite{deng2024cross-scale} and (3) MSPN. Furthermore, we compare with other architectures in related previous studies, including the basic maxpooling and meanpooling methods; graph-based methods of Patch-GCN~\cite{chen2021patchgcn}, H$^2$-MIL~\cite{yu2022h2mil}, CAMIL~\cite{fourkioti2024camil} and Sm-MIL~\cite{castro-mac2024sm}; transformer-based methods of TransMIL~\cite{shao2021transmil}, HIPT~\cite{chen2022hipt} and HAG-MIL~\cite{xiong2023hagmil}; as well as ZoomMIL~\cite{thandiackal2022zoommil}.

For training configuration, we use NLLSurv loss for prognosis prediction following the previous implementations~\cite{zadeh2021nllsurv,chen2022pancancer} while cross-entropy loss is used for all other tasks. All trainings are performed at 150 maximum epochs with a learning rate of $2\times10^{-4}$ and gradient accumulation of 32, while the AdamW optimiser and cosine decay scheduler with T\_max of 145 are used for optimisation, and early stopping is set for patience in 10 epochs. LinearLR is used for warm-up in the beginning 5 epochs with start factor of 0.1. All baselines are implemented following their original design and all experiments are undertaken on two NVIDIA RTX Pro 6000 GPUs.

For evaluation metrics, we use the area under the curve (AUC) to measure the performance of tumour classification, receptor status prediction, and subtyping, Quadratic Weighted Kappa (QWK) for Gleason grading following the official measurement, while the concordance-index (C-index) is used to measure prognosis prediction performance. We employ five-fold cross-validation, with train:val:test ratio of 3:1:1, for training across all tasks that the training, validation and test sets are independent to each other. For CAMELYON16, all folds are tested on the officially assigned independent test set. For external validation on TCGA-BRCA, performance is reported by ensembling the five-fold models.

\section{Ethics approval and consent to participate}
For SteatoSITE, anonymised tissue was supplied after approval by the National Health Service Research Scotland (NRS) Biorepository Network (Reference: SR1032; 2 August 2018). All other datasets are public and do not involve additional ethical approvals.

\section{Data availability}
The NIHR Breast Cancer dataset can be accessed from the NIHR BioResource. TCGA-RCC, TCGA-KIRC, TCGA-LUAD, TCGA-BLCA and TCGA-BRCA can be accessed from the GDC Portal. CAMELYON16 can be accessed from Grand Challenge - CAMELYON16. PANDA can be accessed from Grand Challenge - The PANDA Challenge. SurGen can be accessed from GigaDB. Access to SteatoSITE is available after approved application (\url{https://regeneration-repair.ed.ac.uk/research/research-resources/steatosite}); the RNAseq from which the THRB annotations were derived are deposited in the European Nucleotide Archive (\url{https://www.ebi.ac.uk/ena; study accession number: PRJEB58625}). Gene expression data are freely available for browsing online at \url{https://shiny.igc.ed.ac.uk/SteatoSITE\_gene\_explorer/}.

\section{Code availability}
The code is available at: \url{https://github.com/franksyng/MSPN}.

\bibliography{sn-bibliography}

\section{Acknowledgements}
This work was supported by Medical Research Scotland (PHD-50538-2022). We also thank Nancy Nan and Zhongchi Xue for revising the manuscript.

\section{Author contributions}
Conceptualisation, S.W., Y.Q., H.B., T.J.K.; Methodology, S.W.; Data Acquisition, S.W., J.A.F., T.J.K.; Data analysis, S.W., J.B., I.P.N., S.P., J.A.F., H.B., T.J.K.; Visualisation, S.W., S.P., J.A.F., T.J.K.; Writing - Original Draft, S.W., Y.Q.; Writing - Review \& Editing, S.W., J.B., I.P.N., S.P., H.B., T.J.K.; Funding acquisition: T.J.K; All authors read and approved the submitted version of the manuscript.

\section{Competing interests}
T.J.K. has served as a consultant or advisory board member for Resolution Therapeutics, Clinnovate Health, HistoIndex, Fibrofind, Kynos Therapeutics, Perspectum, Concept Life Sciences, Servier Laboratories, Taiho Oncology, Roche, and Jazz Pharmaceuticals, and has received speakers’ fees from Servier Laboratories, Jazz Pharmaceuticals, Astrazeneca, HistoIndex, and Incyte Corporation. J.B. and I.P.N. were employed by the company Indica Labs, LCC. No other authors declare any conflicts of interest.

\pagebreak
\section{Figures}

\begin{figure}[h]
\begin{center}
\centerline{\includegraphics[width=\textwidth]{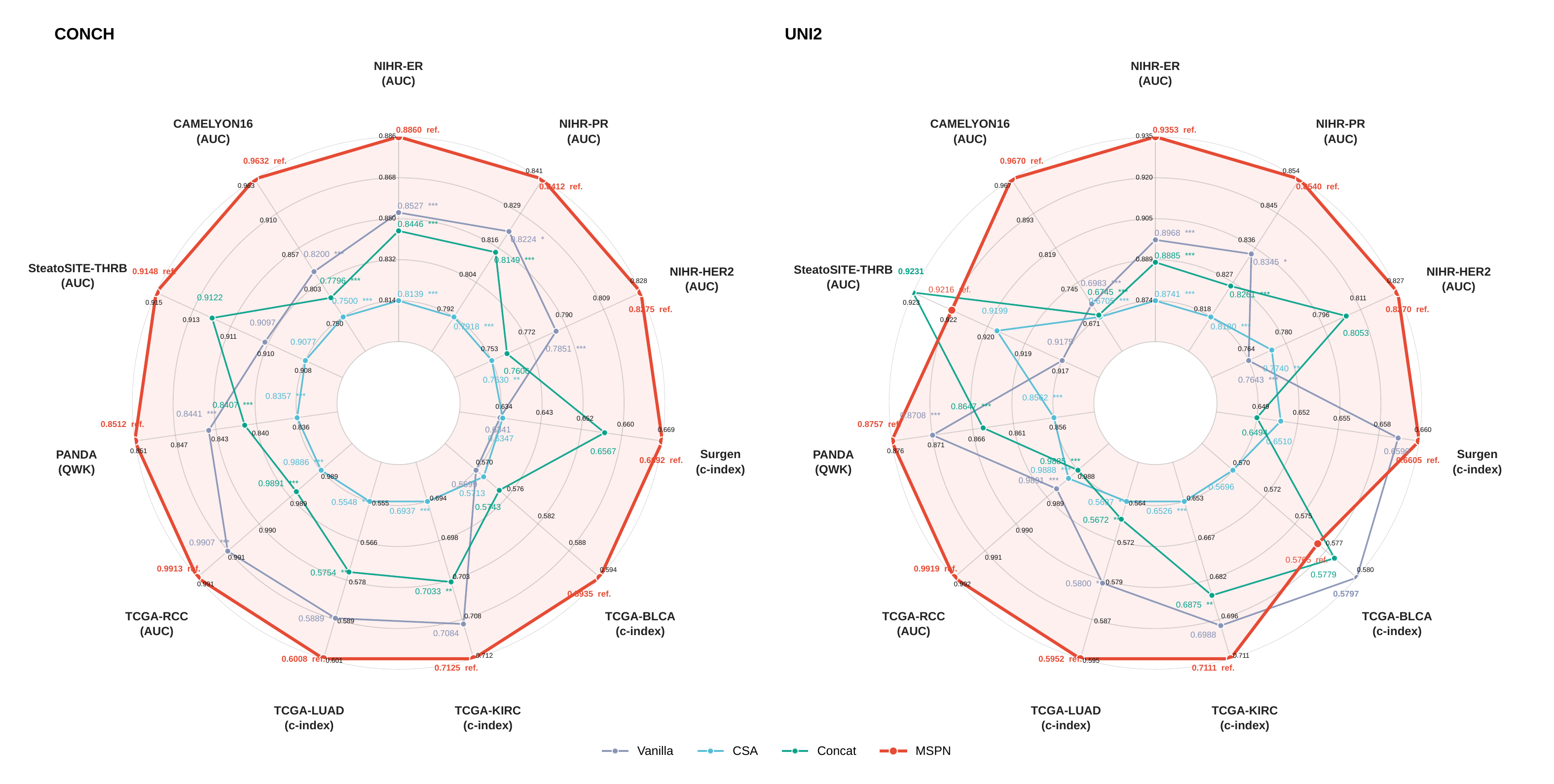}}
\caption{\textbf{MSPN performs consistently better than vanilla models or those with cross-scale attention or concatenation approaches to multi-scale inputs.} AUC score, c-index, and QWK are used to measure performance of classification tasks, prognosis prediction tasks, and Gleason score grading, respectively. The results are reported summarising across ABMIL, DSMIL, CLAM-SB, and CLAM-MB. Mean values are indicated on the radial axes and statistical tests are performed with 2000 paired bootstrap trails, taking MSPN as reference. ***$p<0.001$, **$p<0.01$, *$p<0.05$.}
\label{fig:mspn_vs_attn_radar}
\end{center}
\vskip -0.3in
\end{figure}

\begin{table*}[h]
\centering
\caption{\textbf{MSPN improves performance compared with vanilla MIL formulations and those incorporating mutiple scales with cross-scale attention (CSA) or concatenation (Concat) across 11 tasks.} Mean values of AUC score, c-index, and QWK (measuring classification tasks, prognosis prediction tasks, and Gleason score grading, respectively) across all tasks with standard deviations are reported, and statistical tests are performed with 2000 paired bootstrap trails, taking MSPN as the reference. ***$p<0.001$, **$p<0.01$, *$p<0.05$.}
\label{tab:model_level}
\resizebox{\textwidth}{!}{
\begin{tabular}{l|cccc|cccc}
\toprule
Foundation Model & \multicolumn{4}{c|}{CONCH} & \multicolumn{4}{c}{UNI2} \\
\midrule
Multi-scale & Vanilla & Concat & CSA & MSPN & Vanilla & Concat & CSA & MSPN \\
\midrule
ABMIL & $0.789^{***}_{\pm 0.133}$ & $0.775^{***}_{\pm 0.132}$ & $0.763^{***}_{\pm 0.133}$ & $\mathbf{0.803}_{\pm 0.142}$ & $0.787^{***}_{\pm 0.137}$ & $0.772^{***}_{\pm 0.149}$ & $0.760^{***}_{\pm 0.150}$ & $\mathbf{0.816}_{\pm 0.147}$ \\
DSMIL & $0.771^{***}_{\pm 0.132}$ & $0.774^{***}_{\pm 0.135}$ & $0.757^{***}_{\pm 0.145}$ & $\mathbf{0.806}_{\pm 0.139}$ & $0.773^{***}_{\pm 0.136}$ & $0.777^{***}_{\pm 0.140}$ & $0.783^{***}_{\pm 0.138}$ & $\mathbf{0.809}_{\pm 0.152}$ \\
CLAM-SB & $0.774^{***}_{\pm 0.139}$ & $0.763^{***}_{\pm 0.131}$ & $0.754^{***}_{\pm 0.133}$ & $\mathbf{0.806}_{\pm 0.138}$ & $0.767^{***}_{\pm 0.146}$ & $0.764^{***}_{\pm 0.144}$ & $0.745^{***}_{\pm 0.152}$ & $\mathbf{0.805}_{\pm 0.150}$ \\
CLAM-MB & $0.767^{***}_{\pm 0.138}$ & $0.761^{***}_{\pm 0.132}$ & $0.743^{***}_{\pm 0.131}$ & $\mathbf{0.804}_{\pm 0.142}$ & $0.759^{***}_{\pm 0.146}$ & $0.761^{***}_{\pm 0.145}$ & $0.745^{***}_{\pm 0.143}$ & $\mathbf{0.811}_{\pm 0.152}$ \\
\bottomrule
\end{tabular}
}
\end{table*}

\begin{table*}[h]
\centering
\caption{\textbf{MSPN outperforms the vanilla baseline and multi-scale cross-scale attention (CSA) and concatenation variants of ABMIL, DSMIL, CLAM-SB, and CLAM-MB in external validation on the TCGA-BRCA dataset for ER, PR, and HER2 prediction.} Mean AUC scores with standard deviation are reported and statistical tests are performed with 2000 paired bootstrap trails, taking MSPN as reference. ***$p<0.001$, **$p<0.01$, *$p<0.05$.}
\label{tab:ext_roc}
\resizebox{\textwidth}{!}{
\begin{tabular}{l|l|ccc|ccc}
\toprule
\multicolumn{2}{l|}{Foundation Model} & \multicolumn{3}{c|}{CONCH} & \multicolumn{3}{c}{UNI2} \\
\midrule
Model & Multi-scale & ER ($n=1{,}069$) & PR ($n=1{,}066$) & HER2 ($n=626$) & ER ($n=1{,}069$) & PR ($n=1{,}066$) & HER2 ($n=626$) \\
\midrule
\multirow{4}{*}{ABMIL} & Vanilla & $0.864^{**}_{\pm 0.014}$ & $0.785_{\pm 0.014}$ & $0.629^{***}_{\pm 0.027}$ & $0.881_{\pm 0.064}$ & $0.793_{\pm 0.067}$ & $0.642^{**}_{\pm 0.015}$ \\
 & Concat & $0.845^{***}_{\pm 0.017}$ & $0.759^{***}_{\pm 0.046}$ & $0.647_{\pm 0.006}$ & $\mathbf{0.890}_{\pm 0.005}$ & $0.794_{\pm 0.005}$ & $0.678_{\pm 0.016}$ \\
 & CSA & $0.832^{***}_{\pm 0.014}$ & $0.748^{***}_{\pm 0.027}$ & $0.640_{\pm 0.008}$ & $0.889_{\pm 0.016}$ & $0.796_{\pm 0.006}$ & $0.648_{\pm 0.020}$ \\
 & MSPN & $\mathbf{0.872}_{\pm 0.064}$ & $\mathbf{0.791}_{\pm 0.010}$ & $\mathbf{0.670}_{\pm 0.027}$ & $0.886_{\pm 0.016}$ & $\mathbf{0.798}_{\pm 0.008}$ & $\mathbf{0.687}_{\pm 0.012}$ \\
\midrule
\multirow{4}{*}{DSMIL} & Vanilla & $0.853^{***}_{\pm 0.012}$ & $0.779^{*}_{\pm 0.028}$ & $0.632_{\pm 0.024}$ & $0.877_{\pm 0.012}$ & $0.788_{\pm 0.049}$ & $0.646^{**}_{\pm 0.020}$ \\
 & Concat & $0.862^{*}_{\pm 0.023}$ & $0.780_{\pm 0.052}$ & $\mathbf{0.669}_{\pm 0.020}$ & $\mathbf{0.885}_{\pm 0.009}$ & $0.787_{\pm 0.066}$ & $\mathbf{0.691}_{\pm 0.025}$ \\
 & CSA & $0.853^{**}_{\pm 0.042}$ & $0.779_{\pm 0.017}$ & $0.657_{\pm 0.047}$ & $0.881_{\pm 0.074}$ & $0.788_{\pm 0.015}$ & $0.691_{\pm 0.033}$ \\
 & MSPN & $\mathbf{0.873}_{\pm 0.027}$ & $\mathbf{0.793}_{\pm 0.026}$ & $0.659_{\pm 0.011}$ & $0.875_{\pm 0.019}$ & $\mathbf{0.789}_{\pm 0.026}$ & $0.684_{\pm 0.029}$ \\
\midrule
\multirow{4}{*}{CLAM-SB} & Vanilla & $0.852^{***}_{\pm 0.028}$ & $0.775^{***}_{\pm 0.015}$ & $0.632^{*}_{\pm 0.017}$ & $0.877_{\pm 0.009}$ & $0.789_{\pm 0.003}$ & $0.619^{***}_{\pm 0.022}$ \\
 & Concat & $0.841^{***}_{\pm 0.019}$ & $0.742^{***}_{\pm 0.019}$ & $0.637_{\pm 0.015}$ & $0.875_{\pm 0.132}$ & $0.779^{*}_{\pm 0.007}$ & $0.619^{*}_{\pm 0.019}$ \\
 & CSA & $0.820^{***}_{\pm 0.004}$ & $0.736^{***}_{\pm 0.019}$ & $0.618_{\pm 0.003}$ & $0.836^{***}_{\pm 0.011}$ & $0.758^{***}_{\pm 0.012}$ & $0.582^{**}_{\pm 0.016}$ \\
 & MSPN & $\mathbf{0.871}_{\pm 0.030}$ & $\mathbf{0.793}_{\pm 0.019}$ & $\mathbf{0.663}_{\pm 0.023}$ & $\mathbf{0.884}_{\pm 0.018}$ & $\mathbf{0.799}_{\pm 0.005}$ & $\mathbf{0.684}_{\pm 0.022}$ \\
\midrule
\multirow{4}{*}{CLAM-MB} & Vanilla & $0.845^{***}_{\pm 0.017}$ & $0.763^{***}_{\pm 0.057}$ & $0.616^{***}_{\pm 0.029}$ & $0.871_{\pm 0.059}$ & $0.788_{\pm 0.003}$ & $0.620^{**}_{\pm 0.052}$ \\
 & Concat & $0.830^{***}_{\pm 0.045}$ & $0.748^{***}_{\pm 0.020}$ & $0.639_{\pm 0.011}$ & $0.864^{*}_{\pm 0.006}$ & $0.770_{\pm 0.005}$ & $0.637_{\pm 0.018}$ \\
 & CSA & $0.811^{***}_{\pm 0.047}$ & $0.741^{***}_{\pm 0.086}$ & $0.623_{\pm 0.007}$ & $0.844^{***}_{\pm 0.007}$ & $0.744^{**}_{\pm 0.021}$ & $0.581^{**}_{\pm 0.021}$ \\
 & MSPN & $\mathbf{0.873}_{\pm 0.032}$ & $\mathbf{0.783}_{\pm 0.007}$ & $\mathbf{0.670}_{\pm 0.027}$ & $\mathbf{0.885}_{\pm 0.016}$ & $\mathbf{0.789}_{\pm 0.031}$ & $\mathbf{0.683}_{\pm 0.035}$ \\
\bottomrule
\end{tabular}
}
\end{table*}

\begin{table*}[h]
\centering
\caption{\textbf{MSPN matches or out-performs most graph-based, transformer-based, and multi-magnification paradigms in external validation on the TCGA-BRCA dataset for ER, PR, and HER2 prediction.} The best performing MSPN variant in each task is selected for comparison. Mean AUC scores with standard deviation are reported and statistical tests are performed with 2000 paired bootstrap trails, taking the best MSPN variants as reference. ***$p<0.001$, **$p<0.01$, *$p<0.05$.}
\label{tab:ext_roc_baselines}
\resizebox{\textwidth}{!}{
\begin{tabular}{l|ccc|ccc}
\toprule
Foundation Model & \multicolumn{3}{c|}{CONCH} & \multicolumn{3}{c}{UNI2} \\
\midrule
Method & ER ($n=1{,}069$) & PR ($n=1{,}066$) & HER2 ($n=626$) & ER ($n=1{,}069$) & PR ($n=1{,}066$) & HER2 ($n=626$) \\
\midrule
TransMIL & $0.866_{\pm 0.012}$ & $0.772^{**}_{\pm 0.014}$ & $0.635^{***}_{\pm 0.012}$ & $0.878_{\pm 0.014}$ & $0.787^{*}_{\pm 0.009}$ & $0.664^{*}_{\pm 0.029}$ \\
HAG-MIL & $0.859^{**}_{\pm 0.041}$ & $0.785_{\pm 0.016}$ & $0.632^{**}_{\pm 0.022}$ & $0.873^{*}_{\pm 0.016}$ & $0.791_{\pm 0.011}$ & $0.683_{\pm 0.037}$ \\
CAMIL & $0.855^{***}_{\pm 0.036}$ & $0.770^{**}_{\pm 0.022}$ & $0.638^{***}_{\pm 0.017}$ & $0.878_{\pm 0.019}$ & $0.782^{**}_{\pm 0.016}$ & $0.665^{**}_{\pm 0.021}$ \\
PatchGCN & $0.852^{***}_{\pm 0.014}$ & $0.787_{\pm 0.021}$ & $\mathbf{0.673}_{\pm 0.043}$ & $0.870^{***}_{\pm 0.018}$ & $0.773^{***}_{\pm 0.022}$ & $0.655^{**}_{\pm 0.022}$ \\
HIPT & $0.871_{\pm 0.014}$ & $0.780_{\pm 0.035}$ & $0.660_{\pm 0.028}$ & $0.887_{\pm 0.049}$ & $0.784^{*}_{\pm 0.037}$ & $\mathbf{0.706}_{\pm 0.032}$ \\
H2MIL & $0.855^{***}_{\pm 0.006}$ & $0.784_{\pm 0.018}$ & $0.644^{**}_{\pm 0.008}$ & $0.868^{**}_{\pm 0.036}$ & $0.751^{***}_{\pm 0.028}$ & $0.634^{**}_{\pm 0.061}$ \\
SmMIL & $0.863^{**}_{\pm 0.014}$ & $0.782_{\pm 0.019}$ & $0.636^{***}_{\pm 0.016}$ & $0.885_{\pm 0.013}$ & $0.792^{*}_{\pm 0.009}$ & $0.648^{*}_{\pm 0.020}$ \\
ZoomMIL & $0.855^{**}_{\pm 0.037}$ & $0.761^{**}_{\pm 0.018}$ & $0.648_{\pm 0.031}$ & $\mathbf{0.887}_{\pm 0.015}$ & $\mathbf{0.800}_{\pm 0.029}$ & $0.684_{\pm 0.011}$ \\
\midrule
MSPN (best variant) & $\mathbf{0.873}_{\pm 0.027}$ & $\mathbf{0.793}_{\pm 0.026}$ & $0.670_{\pm 0.027}$ & $0.886_{\pm 0.016}$ & $0.799_{\pm 0.005}$ & $0.687_{\pm 0.012}$ \\
\quad Selected variant & DSMIL & DSMIL & CLAM-MB & ABMIL & CLAM-SB & ABMIL \\
\bottomrule
\end{tabular}
}
\end{table*}

\beginsupplement

\begin{figure}[h]
\begin{center}
\centerline{\includegraphics[width=\textwidth]{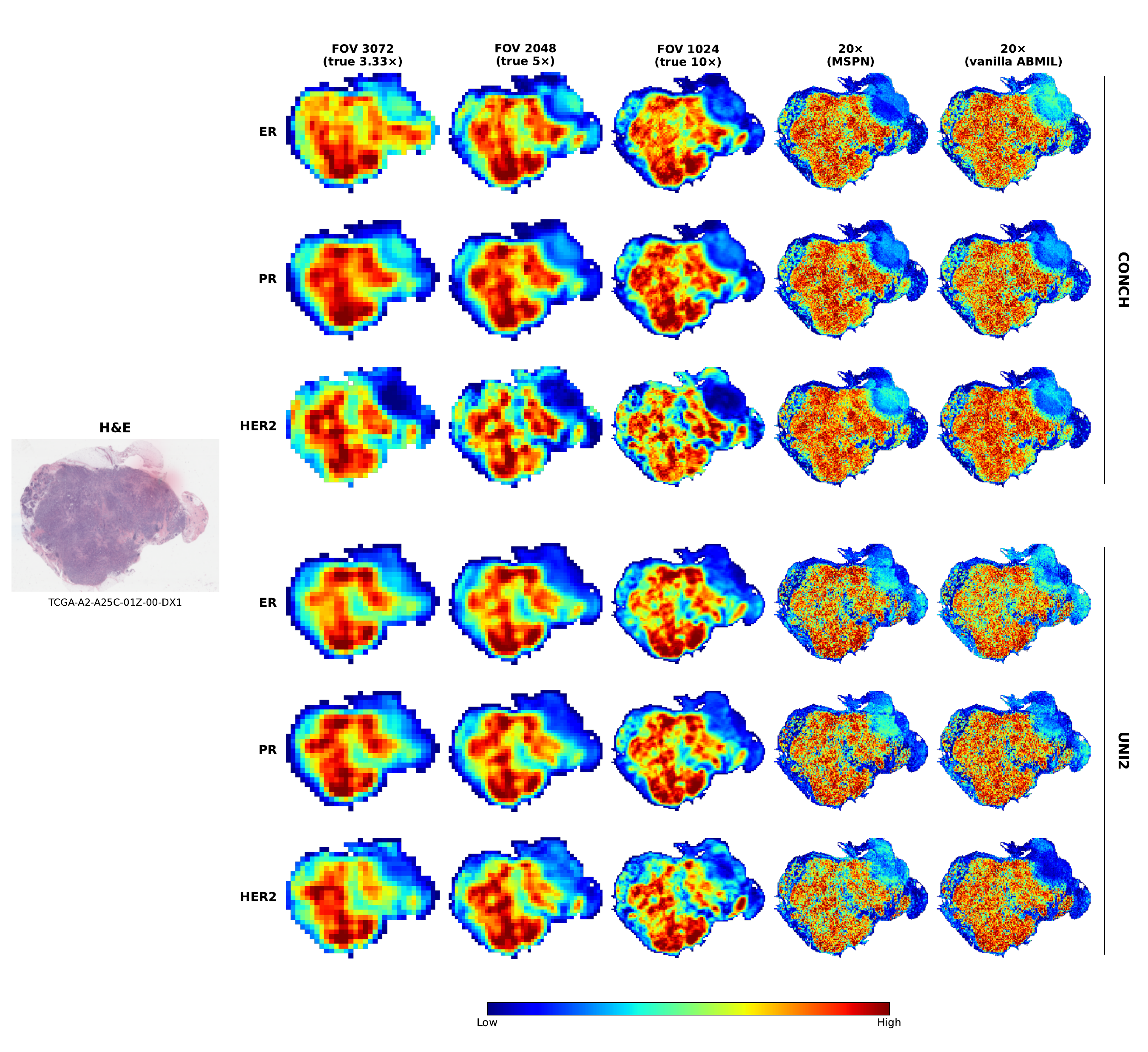}}
\caption{\textbf{Examples of attention heatmap visualisations for ER, PR and HER2 prediction on a case from the TCGA-BRCA external validation task.} Attention heatmaps from the coarse guidances from CGNs with FOV of 3072, 2048 and 1024 in ABMIL+MSPN are shown separately for prediction of ER, PR, and HER2 status using both CONCH and UNI2 as feature extractors, indicating a pattern of progressive focus on relevant areas.}
\label{supfig:vis_simple}
\end{center}
\vskip -0.3in
\end{figure}

\begin{figure}[h]
\begin{center}
\centerline{\includegraphics[width=1.2\textwidth]{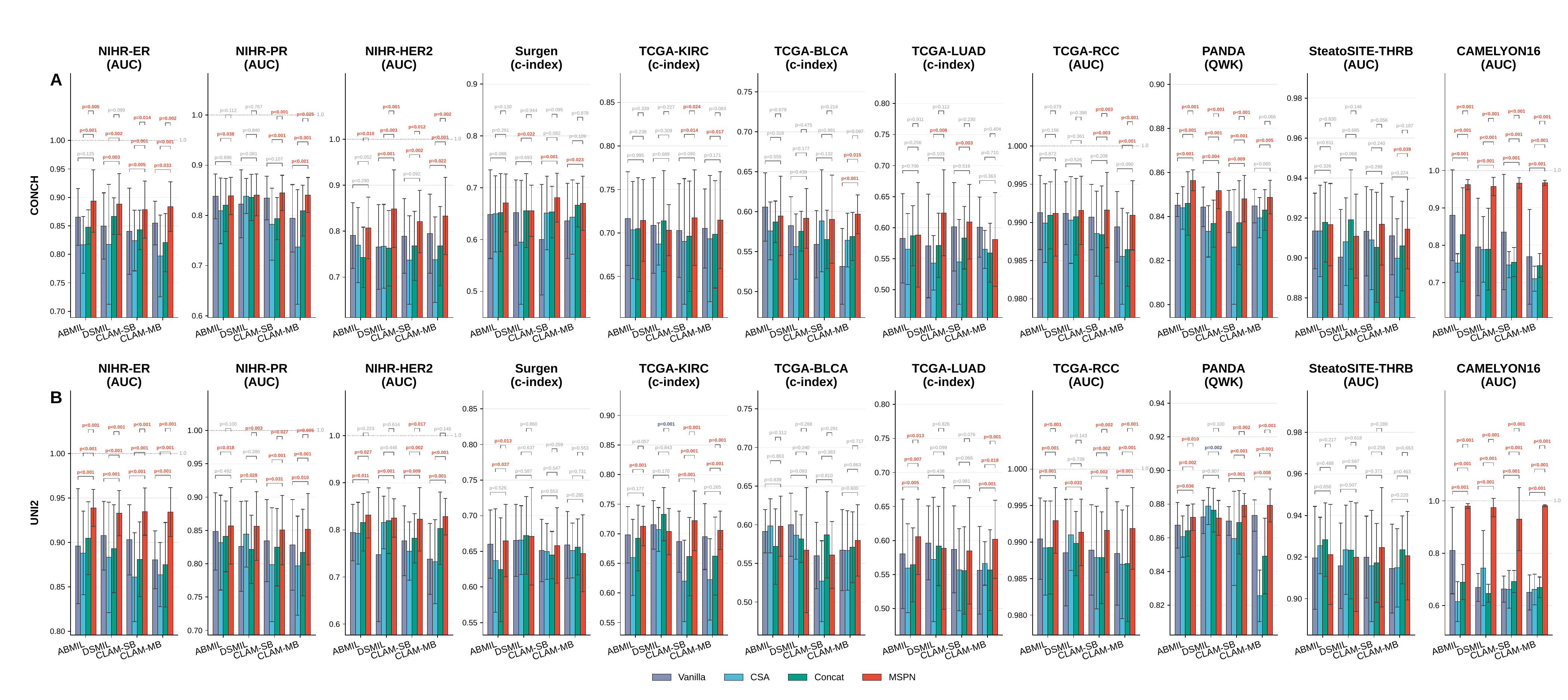}}
\caption{\textbf{Task-specific performance comparisons for vanilla, CSA, concatenation and MSPN.} The results are reported in detailed. Mean values are reported and statistical tests are performed with 2000 paired bootstrap trails, taking MSPN as reference. $P$-values are shown in the figure.}
\label{supfig:mspn_vs_attn_bar}
\end{center}
\vskip -0.3in
\end{figure}

\end{document}